\documentclass[twoside,11pt]{article}

\pdfoutput=1

\usepackage{jmlr2e}

\usepackage{amsmath,amssymb}
\usepackage{epstopdf}
\usepackage{multirow}
\usepackage{array}
\usepackage{subfigure}
\newcolumntype{M}[1]{>{\centering\arraybackslash}m{#1}}
\newcolumntype{X}{%
  >{\rowstyle{\relax}}l%
}
\newcolumntype{Y}{%
  >{\currentrowstyle}S[detect-weight]%
}
\newcommand{\rowstyle}[1]{%
  \protected\gdef\currentrowstyle{#1}%
}

\usepackage{algorithm}
\usepackage{algorithmic}
\usepackage{overpic}
\usepackage{subfigure}
\usepackage{amsmath,amssymb} 
\usepackage{multirow}
\usepackage{overpic}
\usepackage{color}

\firstpageno{1}

\input{def1.set}

\begin{document}

\title{Maximal Sparsity with Deep Networks?}

\author{\name Bo Xin, Yizhou Wang, and Wen Gao \\
       \addr School of Electronics Engineering and Computer Science \\
       Peking University\\
       Beijing, China \\
       Email: \{boxin,yizhou.wang,wgao\}@pku.edu.cn \\
       \AND
       \name David Wipf  \\
       \addr Microsoft Research\\
       Beijing, China \\
       Email: davidwipf@gmail.com
       }

\maketitle

\begin{abstract}
The iterations of many sparse estimation algorithms are comprised of a fixed linear filter cascaded with a thresholding nonlinearity, which collectively resemble a typical neural network layer.  Consequently, a lengthy sequence of algorithm iterations can be viewed as a deep network with shared, hand-crafted layer weights.  It is therefore quite natural to examine the degree to which a learned network model might act as a viable surrogate for traditional sparse estimation in domains where ample training data is available.  While the possibility of a reduced computational budget is readily apparent when a ceiling is imposed on the number of layers, our work primarily focuses on estimation accuracy.  In particular, it is well-known that when a signal dictionary has coherent columns, as quantified by a large RIP constant, then most tractable iterative algorithms are unable to find maximally sparse representations.  In contrast, we demonstrate both theoretically and empirically the potential for a trained deep network to recover minimal $\ell_0$-norm representations in regimes where existing methods fail.  The resulting system is deployed on a practical photometric stereo estimation problem, where the goal is to remove sparse outliers that can disrupt the estimation of surface normals from a 3D scene.

\end{abstract}


\section{Introduction}
Our launching point is the optimization problem
\begin{equation} \label{eq:basic_L0_prob}
\min_{\bx} \| \bx \|_0 ~~~ \mbox{s.t. } \by = \bPhi \bx,
\end{equation}
where $\by \in \mathbb{R}^{n}$ is an observed vector, $\bPhi \in \mathbb{R}^{n \times m}$ is some known, overcomplete dictionary of feature/basis vectors with $m>n$, and $\| \cdot \|_0$ denotes the $\ell_0$ norm of a vector, or a count of the number of nonzero elements.  Consequently, (\ref{eq:basic_L0_prob}) can be viewed as the search for a maximally sparse vector $\bx^*$ such that $\by$ can be represented using the fewest number of features in the feasible region.\footnote{In practice, it is common to relax the feasible region to $\| \by - \bPhi \bx\|_2 \leq \epsilon$, or replace the constraint altogether with a sensible data fit term balanced with a trade-off parameter.}  Unfortunately however, direct assault on (\ref{eq:basic_L0_prob}) involves an intractable, combinatorial optimization process, and therefore efficient alternatives that return a maximally sparse $\bx^*$ with high probability in restricted regimes are sought.  Popular examples with varying degrees of computational overhead include convex relaxations such as $\ell_1$-norm minimization \citep{Donoho03,Tibshirani96}, greedy approaches like orthogonal matching pursuit (OMP) \citep{Pati93,Tropp04}, and many flavors of iterative thresholding \citep{Beck09,Blumensath08}.

Variants of these algorithms find practical relevance in numerous disparate application domains, including feature selection \citep{Cotter02,Figueiredo01}, outlier removal \citep{Candes05,Ikehata12}, compressive sensing \citep{Donoho06b}, and source localization \citep{Baillet01,Malioutov05} among many others.  However, a fundamental weakness underlies them all: If the Gram matrix $\bPhi^{\top} \bPhi$ has significant off-diagonal energy, indicative of strong coherence between columns of $\bPhi$, then estimation of $\bx^*$ may be extremely poor.  Indeed both the cardinality of the solution, and often more importantly, the locations of nonzero elements, can be completely suboptimal.  Loosely speaking this occurs because, as higher correlation levels are present, the null-space of $\bPhi$ is more likely to include large numbers of approximately sparse vectors that tend to distract existing algorithms in the feasible region.  The degree to which this risk is present can be quantified by a so-called \emph{restricted isometry} constant to be described in detail later.  Compounding the problem is that, in all but the most ideal settings where we are free to choose $\bPhi$ randomly from certain favorable distributions, there is no way of knowing in advance the true degree in which this correlation structure will be disruptive (e.g., restricted isometry constants are actually not feasible to compute in practice).

In this paper we consider recent developments in the field of deep learning as an entry point for improving the performance of sparse recovery algorithms.  Although seemingly unrelated at first glance, the layers of a deep neural network (DNN) can be viewed as iterations of some algorithm that have been unfolded into a network structure \citep{Gregor10,Hershey14}.  In particular, iterative thresholding approaches such as those mentioned above typically involve an update rule  comprised of a fixed, linear filter followed by a non-linear activation function that promotes sparsity.  Consequently, algorithm execution can be interpreted as passing an input through an extremely deep network with constant layer weights (dependent on $\bPhi$) at every layer.

This `unfolding' viewpoint immediately suggests that we consider substituting discriminatively learned weights in place of those inspired by the original sparse recovery algorithm.  For example, it has been argued that, given access to a sufficient number of $\{\bx^*,\by\}$ pairs, a trained network may be capable of producing quality sparse estimates with a modest number of layers.  This in turn can lead to a dramatically reduced computational burden relative to purely optimization-based approaches, which can require hundreds or even thousands of iterations to sufficiently converge \citep{Gregor10,Sprechmann15}.

Existing work on sparse estimation through deep network training borrows basic network components directly from the underlying iterative algorithm.  Different networks are primarily differentiated by the types of activation functions employed, which performed as sparsity-promoting non-linearities during their former life in service to iterative optimization.  For example, \citep{Gregor10} promotes a soft-threshold function inspired by and iterative shrinkage-thresholding algorithm (ISTA) for minimizing the $\ell_1$-norm, a well-known convex approximation to the canonical $\ell_0$ norm sparsity penalty from (\ref{eq:basic_L0_prob}).  In contrast, \citep{Sprechmann15} advocates a wider class of functions derived from proximal operators \citep{Parikh14}.  Finally, it has also been suggested that replacing typically continuous activation functions with hard-threshold operators may lead to sparser representations \citep{Wang15l0}.  At a high level though, one common ingredient of all of these approaches is the adoption of shared weights across layers.

While existing empirical results are promising, especially in terms of the reduction in computational footprint, there is as of yet no empirical demonstration of a learned deep network that can unequivocally recover maximally sparse vectors $\bx^*$ with greater accuracy than conventional, state-of-the-art optimization-based algorithms.  Nor is there supporting theoretical evidence elucidating the exact mechanism whereby learning may be expected to improve the estimation accuracy, especially in the presence of coherent dictionaries $\bPhi$.  Additionally, minimal insights exist that might be transferrable to assessing the behavior of broader learning objectives and systems.

\subsection{Paper Overview}

This paper attempts to fill in these gaps described above, at least to the extent possible, via the following organizational structure.  In Section \ref{sec:IHT_to_DNN} we begin by reviewing the iterative hard-thresholding (IHT) algorithm for estimating a sparse vector.  IHT was chosen because it can be directly unfolded for learning purposes, is representative of many sparse estimation paradigms, and is amenable to theoretical analysis.  Next we discuss the limitations of IHT, including its high sensitivity to correlated designs, and motivate a DNN-like, unfolded alternative.  Later, Section \ref{sec:constant_weights} considers this unfolded IHT network with shared layer-wise weights and activations, which is the standard template for existing methods.  We explicitly quantify the degree to which such networks can compensate for correlations in $\bPhi$, but also expose the breaking point whereby any possible shared-weight construction is likely to fail.

This naturally segues to richer deep networks with layer-wise independent weights and activations (meaning different layers need not share the same, fixed weights and activations), which we scrutinize in Section \ref{sec:adaptive_weights}.  Here we describe a multi-resolution dictionary, with highly correlated clusters of columns, that explicitly requires the richer class of layer parameterizations to guarantee successful sparse recovery.  Section \ref{sec:multi_resolution} then further elucidates the essential multi-resolution nature of the sparse estimation problem, and how we may deviate from strict adherence to any particular unfolded algorithmic script in designing a practical DNN.  In particular, we motivate a multi-label classification network to focus on learning correct support patterns.  In Section \ref{sec:training} we describe what we believe to be an essential ingredient for constructing an effective training set.  Corroborating simulation results and a real-world computer vision application example are presented in Sections \ref{sec:simulations} and \ref{sec:application} respectively, followed by exploration of alternative recurrent long-short-term-memory (LSTM) structures in Section \ref{sec:lstm}.  Final discussions are in Section \ref{sec:discussion}, while all proofs are deferred to the Appendix.

\subsection{Summary of Contributions}

Our technical and empirical contributions can be distilled to the following points:
\begin{itemize}
\item We rigorously dissect the benefits of unfolding conventional sparse estimation algorithms to produce trainable deep networks.  This includes a precise characterization of exactly how different architecture choices can affect the ability to improve effective restrictive isometry constants, which measure of the degree of disruptive correlation present in a dictionary.  This helps to quantify the limits of shared layer weights and motivates more flexible network constructions that account for multi-resolution structure in $\bPhi$ in a previously unexplored fashion.  Importantly, we envision that our analyses are emblematic of important factors present in other DNN-related domains.

\item Based on these theoretical insights, and a better understanding of the essential factors governing performance, we establish the degree to which it is favorable to diverge from strict conformity to any particular unfolded algorithmic script.  In particular, we argue that the equivalent of layer-wise independent weights and/or activations are essential, while retainment of original hard-thresholding non-linearities and squared-error loss implicit to IHT and related algorithms is not.  We also recast the the core problem as deep multi-label classification given that optimal support pattern is our primary concern.  This allows us to adopt a novel training paradigm that is less sensitive to the specific distribution encountered during testing.  Ultimately, we development the first, ultra-fast sparse estimation algorithm that can effectively deal with coherent dictionaries and adversarial restricted isometry constants.

\item We apply the proposed system to a practical photometric stereo computer vision problem, where the goal is to estimate the 3D geometry of an object using only 2D photos taken from a single camera under different lighting conditions.  In this context, shadows and specularities represent sparse outliers that must be simultaneously removed from $\sim 10^4-10^6$ surface points.  We achieve state-of-the-art performance despite a minuscule computational budget appropriate for real-time mobile environments.

    \item Finally, we explore the connection between unfolded sparse estimation algorithms and unfolded recurrent LSTM networks, revealing that the gating functions intrinsic to the latter can improve performance in the former by allowing coarse-resolution sparsity patterns to prorogate to deeper layers.

\end{itemize}

\section{From Iterative Hard Thesholding (IHT) to Deep Neural Networks} \label{sec:IHT_to_DNN}

This section first introduces IHT before detailing its unfolded DNN analogue.

\subsection{Introduction to IHT}

With knowledge of an upper bound on the true cardinality, solving (\ref{eq:basic_L0_prob}) can be replaced by the equivalent problem
\begin{equation} \label{eq:basic_IHT_problem}
\min_{\bx} \tfrac{1}{2} \| \by - \bPhi \bx \|_2^2    ~~~ \mbox{s.t. } \| \bx \|_0 \leq k.
\end{equation}
Iterative hard-thresholding (IHT) attempts to minimize (\ref{eq:basic_IHT_problem}) using what can be viewed as computationally-efficient projected gradient iterations \citep{Blumensath09}.  Let $\bx^{(t)}$ denote the estimate of some maximally sparse $\bx^*$ after $t$ iterations.  IHT first computes the gradient of the quadratic objective evaluated at $\bx^{(t)}$ given by
\begin{equation}
\left. \nabla_{\bx} \right|_{\bx = \bx^{(t)}} = \bPhi^{\top} \bPhi \bx^{(t)} - \bPhi^{\top} \by.
\end{equation}
We then take the unconstrained gradient step
\begin{equation}
\bx^{(t+1)} = \bx^{(t)} - \mu \left. \nabla_{\bx} \right|_{\bx = \bx^{(t)}},
\end{equation}
 where $\mu$  is a step-size parameter.  Finally, we project onto the constraint set by zeroing out all but the $k$ largest values (in magnitude) of $\bx^{(t+1)}$ using a hard-thresholding operator denoted $H_k[\cdot]$.  The combined iteration becomes
\begin{equation} \label{eq:IHT_iteration}
\bx^{(t+1)} = H_k \left[ \bx^{(t)} - \mu \bPhi^{\top} \left( \by - \bPhi \bx^{(t)}  \right) \right],
\end{equation}
which only requires matrix-vector multiples and is computationally cheap to implement.  For the vanilla version of IHT, the step-size $\mu = 1$ leads to a number of recovery guarantees whereby iterating (\ref{eq:IHT_iteration}), starting from $\bx^{(0)} = {\bf 0}$ is guaranteed to reduce (\ref{eq:basic_IHT_problem}) at each step before eventually converging to the globally optimal solution.\footnote{Other values of $\mu$ or even a positive definite matrix, adaptively chosen, can lead to a faster convergence rate \citep{Blumensath10}.}  These results hinge on properties of $\bPhi$ which relate to the coherence structure of dictionary columns as encapsulated by the following definition.
\begin{definition}[Restricted Isometry Property]
A dictionary $\bPhi$ satisfies the Restricted Isometry Property (RIP) with constant $\delta_k[\bPhi] < 1$ if
\begin{equation}
(1-\delta_k[\bPhi] ) \| \bx \|_2^2 \leq \| \bPhi \bx \|_2^2 \leq (1+\delta_k[\bPhi]) \| \bx \|_2^2
\end{equation}
holds for all $\{\bx : \| \bx \|_0 \leq k\}$. 
\end{definition}
In brief, the smaller the value of the restricted isometry constant $\delta_k[\bPhi]$, the closer any sub-matrix of $\bPhi$ with $k$ columns is to being orthogonal (i.e., it has less correlation structure).

It is now well-established that dictionaries with smaller values of $\delta_k[\bPhi] $ lead to sparse recovery problems that are inherently easier to solve.  In the context of IHT, it has been shown \citep{Blumensath09} that if $\by = \bPhi \bx^*$, with $\| \bx^* \|_0 \leq k$ and $\delta_{3k}[\bPhi] < 1/\sqrt{32}$, then at iteration $t$ of (\ref{eq:IHT_iteration})
\begin{equation}
\|\bx^{(t)} - \bx^*\|_2 \leq 2^{-t}\|\bx^*\|_2.
\end{equation}
It follows that as $t \rightarrow \infty$, $\bx^{(t)} \rightarrow \bx^*$, meaning that we recovery the true, generating $\bx^*$.  Moreover, it can be shown that this $\bx^*$ is also the unique, optimal solution to (\ref{eq:basic_L0_prob}) \citep{Candes06}.

\subsection{Unfolding IHT Iterations}

The success of IHT in recovering maximally sparse solutions crucially depends on the RIP-based condition that $\delta_{3k}[\bPhi] < 1/\sqrt{32}$, which heavily constrains the degree of correlation structure in $\bPhi$ that can be tolerated.  While dictionaries with columns drawn independently and uniformly from the surface of a unit hypersphere\footnote{If elements of $\bPhi$ are drawn iid from $\mathcal{N}(0,1/\sqrt{n})$ and rescaled to have unit $\ell_2$ norm, then the resulting columns will be iid distributed uniformly on the unit sphere.  Moreover, as $n \rightarrow \infty$, each $\ell_2$ column norm converges to one such that normalization is not even necessary.} will satisfy this condition with high probability provided $k$ is small enough \citep{Candes05}, for many/most practical problems of interest we cannot rely on this type of IHT recovery guarantee.  In fact, except for randomized dictionaries in high dimensions where tight bounds exist, we cannot even compute the value of $\delta_{3k}[\bPhi]$, which requires calculating the spectral norm of ${m\choose 3k}$  subsets of dictionary columns.

There are many ways nature might structure a dictionary such that IHT (or most any other existing sparse estimation algorithm) will fail.  Here we consider one of the most straightforward forms of dictionary coherence that can easily disrupt performance.  Consider the situation where $\bPhi = \left[ \epsilon \bA + \bu \bv^{\top} \right] \bN$, where columns of $\bA \in \mathbb{R}^{n\times m}$ and $\bu \in \mathbb{R}^{n}$  are drawn iid from the surface of a unit hypersphere, while $\bv \in \mathbb{R}^{m }$ is arbitrary. Additionally, $\epsilon > 0$ is a scalar and $\bN$ is a diagonal normalization matrix that scales each column of $\bPhi$ to have unit $\ell_2$ norm.  It then follows that if $\epsilon$ is sufficiently small, the rank-one component begins to dominate, and there is no value of $3k$ such that $\delta_{3k}[\bPhi] < 1/\sqrt{32}$.

It is here we hypothesize that DNNs provide a potential avenue for improvement to the extent that they might be able to compensate for disruptive correlation structure in $\bPhi$. To see this, note that from a qualitative standpoint it is quite clear that the iterations of sparsity-promoting algorithms like IHT resemble the layers of neural networks \citep{Gregor10}.  Therefore we can view a long sequence of such iterations as a DNN with fixed, parameterized weights at every layer.  However, what if we are able to learn alternative weights that somehow overcome the limitations of a poor RIP constant?

For example, at the most basic level we might consider general networks with the layer $t+1$ defined by
\begin{equation} \label{eq:general_IHT_layer}
\bx^{(t+1)} = f \left[\bPsi \bx^{(t)} + \bGamma \by \right],
\end{equation}
where $f : \mathbb{R}^{m} \rightarrow \mathbb{R}^{m}$ is a non-linear activation function, and $\bPsi \in \mathbb{R}^{m \times m}$ and $\bGamma \in \mathbb{R}^{m \times n}$ are arbitrary.   Moreover, given access to training pairs $\{\bx^*,\by \}$, where $\bx^*$ is a sparse vector such that $\by = \bPhi \bx^*$, we can optimize $\bPsi$ and $\bGamma$ using traditional stochastic gradient descent just like any other DNN structure.  In the next section we will precisely characterize the extent to which this modification affords any benefit over IHT using $f(\cdot) = H_k[\cdot]$.  Later in Section \ref{sec:adaptive_weights} we will consider adaptive non-linearities $f^{(t)}$ and layer-specific parameters $\{\bPsi^{(t)}, \bGamma^{(t)}\}$.

\section{Analysis using Shared Layer-Wise Weights and Activations} \label{sec:constant_weights}
For simplicity in this section we restrict ourselves to the fixed hard-threshold operator $H_k[\cdot]$ across all layers; however, many of the conclusions borne out of our analysis nonetheless carry over to a much wider range of activation functions $f$.  In general it is difficult to analyze how arbitrary $\bPsi$ and $\bGamma$ may improve upon the fixed parameterization from (\ref{eq:IHT_iteration}) where $\bPsi = \bI - \bPhi^{\top} \bPhi$ and $\bGamma =  \bPhi^{\top}$ (assuming $\mu = 1$).  Fortunately though, we can significantly collapse the space of potential weight matrices by including the natural requirement that if $\bx^*$ represents the true, maximally sparse solution, then it must be a fixed-point of $(\ref{eq:general_IHT_layer})$.  Indeed, without this stipulation the iterations could diverge away from the globally optimal value of $\bx$, something IHT itself will never do.  These considerations lead to the following:

\begin{proposition} \label{prop:constrained_IHT_layer}
Consider a generalized IHT-based network layer given by
\begin{equation} \label{eq:general_IHT_layer2}
\bx^{(t+1)} = H_k \left[\bPsi \bx^{(t)} + \bGamma \by \right]
\end{equation}
and let $\bx^*$ denote any unique, maximally sparse feasible solution to $\by = \bPhi \bx$ with $\| \bx \|_0 \leq k$.  Then to ensure that any such $\bx^*$ is a fixed point of (\ref{eq:general_IHT_layer2}) it must be that $\bPsi = \bI -\bGamma \bPhi$.
\end{proposition}

Although $\bGamma$ remains unconstrained, this result has restricted $\bPsi$ to be a rank-$n$ factor, parameterized by $\bGamma$, subtracted from an identity matrix.  Certainly this represents a significant contraction of the space of `reasonable' parameterizations for a general IHT layer.  In light of Proposition \ref{prop:constrained_IHT_layer}, we may then further consider whether the added generality of $\bGamma$ (as opposed to the original fixed assignment $\bGamma = \bPhi^{\top}$)  affords any further benefit to the revised IHT update
\begin{equation} \label{eq:constrained_IHT_layer}
\bx^{(t+1)} = H_k \left[ \left(\bI -\bGamma \bPhi \right) \bx^{(t)} + \bGamma \by \right].
\end{equation}
For this purpose we note that (\ref{eq:constrained_IHT_layer}) can be interpreted as a projected gradient descent step for solving
\begin{equation}
\min_{\bx} \tfrac{1}{2} \bx^{\top} \bGamma \bPhi \bx - \bx^{\top} \bGamma \by    ~~~ \mbox{s.t. } \| \bx \|_0 \leq k.
\end{equation}
However, if $\bGamma \bPhi$ is not positive semi-definite, then this objective is no longer even convex, and combined with the non-convex constraint is likely to produce an even wider constellation of troublesome local minima with no clear affiliation with the global optimum of our original problem from (\ref{eq:basic_IHT_problem}).  Consequently it does not immediately appear that $\bGamma \neq \bPhi^{\top}$ is likely to provide any tangible benefit.  However, there do exist important exceptions.

The first indication of how learning a general $\bGamma$ might help comes from the following result:
\begin{proposition} \label{prop:constant_layer_improvement}
Suppose that $\bGamma = \bD \bPhi^{\top} \bW \bW^{\top}$, where $\bW$ is an arbitrary matrix of appropriate dimension and $\bD$ is a full-rank diagonal that jointly solve
\begin{equation} \label{eq:delta_optimum}
\delta^*_{3k}\left[\bPhi \right] \triangleq \inf_{\bW,\bD} \delta_{3k} \left[\bW \bPhi \bD \right]. 
\end{equation}
Moreover, assume that $\bPhi$ is substituted with $\bPhi D$ in (\ref{eq:constrained_IHT_layer}), meaning we have simply replaced $\bPhi$ with a new dictionary that has scaled columns.  Given these qualifications, if $\by = \bPhi \bx^*$, with $\| \bx^* \|_0 \leq k$ and $\delta^*_{3k}\left[\bPhi \right] < 1/\sqrt{32}$, then at iteration $t$ of (\ref{eq:constrained_IHT_layer})
\begin{equation}
\|\bD^{-1} \bx^{(t)} - \bD^{-1} \bx^*\|_2 \leq 2^{-t}\|\bD^{-1} \bx^*\|_2.
\end{equation}
\end{proposition}
As before, it follows that as $t \rightarrow \infty$, $\bx^{(t)} \rightarrow \bx^*$, meaning that we recovery the true, generating $\bx^*$.  Additionally, it can be guaranteed that after a finite number of iterations, the correct support pattern will be discovered.  And it should be emphasized that rescaling $\bPhi$ by some known diagonal $\bD$ is a common prescription for sparse estimation (e.g., column normalization) that does not alter the optimal $\ell_0$-norm support pattern.\footnote{Inclusion of this diagonal factor $\bD$ can be equivalently viewed as relaxing Proposition \ref{prop:constrained_IHT_layer} to hold under some fixed rescaling of $\bPhi$, i.e., the optimal support pattern is preserved.}

But the real advantage over regular IHT comes from the fact that $\delta^*_{3k}\left[\bPhi \right] \leq \delta_k \left[ \bPhi \right]$, and in many practical cases, $\delta^*_{3k}\left[\bPhi \right] \ll \delta_{3k} \left[ \bPhi \right]$, which implies success can be guaranteed across a much wider range of RIP conditions.  For example, if we revisit the dictionary $\bPhi = \left[ \epsilon \bA + \bu \bv^{\top} \right] \bN $, an immediate benefit can be observed.  More concretely, for $\epsilon$ sufficiently small we argued that $\delta_{3k} \left[\bPhi \right] > 1/\sqrt{32}$ for all $k$, and consequently convergence to the optimal solution may fail.  In contrast, it can be shown that $\delta^*_{3k}\left[\bPhi \right]$ will remain quite small, satisfying $\delta^*_{3k}\left[\bPhi \right] \approx \delta_{3k} \left[\bA \right]$,  implying  that performance will nearly match that of an equivalent recovery problem using $\bA$ (and as we discussed above, $\delta_{3k} \left[\bA \right]$ is likely to be relatively small per its unique, randomized design).  The following result generalizes a sufficient regime whereby this is possible:
\begin{corollary} \label{cor:constant_layer_improvement}
Suppose $\bPhi = \left[ \epsilon \bA + \bDelta_r \right] \bN$, where elements of $\bA$ are drawn iid from $\mathcal{N}(0,1/\sqrt{n})$, $\bDelta_r$ is any arbitrary matrix with $\mbox{rank}[\bDelta_r] = r < n$, and $\bN$ is a diagonal matrix that enforces unit $\ell_2$ column norms.  Then
\begin{equation}
\mbox{E}\left( \delta^*_{3k}\left[\bPhi \right] \right) \leq \mbox{E}\left( \delta_{3k}\left[ \widetilde{\bA} \right] \right),
\end{equation}
where $\widetilde{\bA}$ denotes the matrix $\bA$ with any $r$ rows removed.
\end{corollary}

Additionally, as the size of $\bPhi$ grows proportionally larger, it can be shown that with overwhelming probability $\delta^*_{3k}\left[\bPhi \right] \leq \delta_{3k}\left[  \widetilde{\bA} \right]$.  Overall, these results suggest that we can essentially annihilate any potentially disruptive rank-$r$  component $\bDelta_r$ at the cost of implicitly losing $r$ measurements (linearly independent rows of $\bA$, and implicitly the corresponding elements of $\by$).  Therefore, at least provided that $r$ is sufficiently small such that $\delta_{3k}\left[  \widetilde{\bA} \right] \approx \delta_{3k}\left[  \bA \right]$, we can indeed be confident that a modified form of IHT can perform much like a system with an ideal RIP constant.\footnote{Of course at some point we will experience diminishing marginal returns using this prescription.  For example, in the extreme case if $r = n-1$, then columns of $\widetilde{\bA}$ will be reduced to a $n-r = 1$ dimensional subspace, and no RIP conditions can possibly hold (see \citep{Bah10} for details of how the RIP constant scales with the dimensions of Gaussian iid matrices).  Regardless, we can still choose some alternative $\bW$ and $\bD$ such that (\ref{eq:delta_optimum}) is optimal, but the optimal solution will no longer involve complete eradication of $\bDelta_r$.}  And of course in practice we may not ever be aware exactly how the dictionary decomposes as some $\bPhi \approx \left[ \epsilon \bA + \bDelta_r \right] \bN$; however, to the extent that this approximation can possibly hold, the effective RIP constant can be improved nonetheless.

It should be noted that globally solving (\ref{eq:delta_optimum}) is non-differentiable and intractable, but this is the whole point of incorporating a DNN network to begin with.  If we have access to a large number of training pairs $\{\bx^*,\by \}$ generated using the true $\bPhi$, then during the course of the learning process a useful $\bW$ and $\bD$ can be implicitly learned such that a maximal number of sparse vectors can be successfully recovered.

Moving forward beyond RIP-related issues, there exists one additional way that learning $\bGamma$ could afford some value.  Suppose now that $\bGamma = \bB \bB^{\top} \bD \bPhi^{\top} \bW \bW^{\top}$, where $\bW$ and $\bD$ are as before (preserving their attendant benefits) and $\bB$ is an arbitrary invertible matrix.  Given that multiplying a gradient by a positive-definite, symmetric matrix guarantees a descent direction is preserved, the inclusion of $\bB \bB^{\top}$ could be viewed as as a form of natural gradient direction to be learned during training \citep{Amari98}.  However, given that such a direction must be universal across all layers and possible sparsity patterns, unlike the universal benefit of a lowered RIP constant, it is unclear the extent to which this $\bB \bB^{\top}$ improves performance.  It would be interesting to isolate this effect at least empirically, but we do not pursue this issue further herein.

To summarize then, learning layer-wise fixed weights $\bPsi$ and $\bGamma$ can indeed provide an important benefit by implicitly reducing the RIP constant of $\bPhi$.  We believe this to be a practically-realizable way of affecting what is otherwise an NP-hard constant to even compute, let alone optimize. Learning layer-wise fixed weights can also produce an alternative `natural gradient' direction; however, given that this direction must be the same for all layers and for all sparse vectors $\bx^*$, it remains unclear whether or not this latter capability provides any tangible welfare.

\section{Analysis using Layer-Wise Independent Weights and Activations } \label{sec:adaptive_weights}
\label{sec:intr}

In the previous section we observed how jointly adjusting $\bW$ and $\bD$ could implicitly remove the effects of low-rank components that inflate dictionary coherence and RIP constant values.  However, we also qualified the advantages of this strategy, with diminishing marginal returns as more non-ideal components enter the picture.  In fact, it is not difficult to describe a slightly more sophisticated scenario such that use of layer-wise constant weights and activations are no longer capable of lowering $\delta_{3k}[\bPhi]$ at all, portending failure when it comes to accurate sparse recovery.  In contrast, this section will reveal that independent weights and adaptive activations can nonetheless still succeed.

To illustrate this effect, we will now analyze dictionaries with columns that are tightly grouped into clusters such that the within-group correlation is high while the between-group correlation is modest.  As further technical results require a bit more precision, we first present the following formal definition.

\begin{definition}[Clustered Dictionary Model]
Let $\bA = [\bA_1, \ldots, \bA_c] \in \mathbb{R}^{n \times m}$ denote a partitioned matrix, with each partition $\bA_j \in \mathbb{R}^{n \times m_j}$ sized such that $m = \sum_i m_i$.  Also define $\bU = [\bu_1,\ldots,\bu_c] \in \mathbb{R}^{n\times c}$ and $\bv_j \in \mathbb{R}^{m_j}$ for all $j = 1,\ldots,c$.   Moreover, assume that both $\bU$ and $\bA$ are constructed with columns of unit $\ell_2$ norm.  Then a dictionary matrix $\bPhi$ is said to arise from the \emph{clustered dictionary model} if $\bPhi = [\bPhi_1,\ldots,\bPhi_c] \bN$, with $\bPhi_j = \bu_j \bv_j^{\top} + \epsilon \bA_j$, $\epsilon > 0$ as a scalar weighting factor, and $\bN$ a diagonal matrix that applies final $\ell_2$ column normalization.  We also define the cluster support $\mathcal{S}_c(\bx) \subset \{1,\ldots,c\}$ as the set of cluster indices whereby some $\bx \in \mathbb{R}^{m}$ has at least one nonzero corresponding element.
\end{definition}

Therefore it becomes readily apparent that, provided $\epsilon$ is chosen sufficiently small, each partition $\bPhi_j$ will be a tight cluster of basis vectors centered around an axis formed by the corresponding $\bu_j$. In some sense this model represents the simplest partitioning of correlation structure into two scales: the inter- and intra-cluster structures.  It thus represents an accessible model for evaluating further manual or learned modifications of IHT.  In particular, we note that assuming $c$ is large, possibly even larger than $n$, we can no longer rely on $\bW$ and $\bD$ to reduce $\delta_{3k}[\bPhi]$ as we did in Section \ref{sec:constant_weights}, as annihilating every rank-one $\bu_j \bv_j^{\top}$ term is clearly impossible.

We now turn to an adaptation of IHT that includes two important modifications that are reflective of many generic DNN structures:
\begin{enumerate}
\item The hard-thresholding operator is generalized to account for prior information about learned support patterns from previous iterations, and
\item We allow the dictionary or weight matrix to change from iteration to iteration sequencing through a fixed set akin to layers of a DNN.
\end{enumerate}
Regarding the former, we define an IHT iteration with partially known support as
\begin{equation} \label{eq:general_IHT_layer2b}
\bx^{(t+1)} = H_{k^{(t)}}\left[\bPsi \bx^{(t)} + \bGamma \by ; \Omega_{\tiny \mbox{on}}^{(t)},\Omega_{\tiny \mbox{off}}^{(t)}  \right],
\end{equation}
where $\Omega_{\tiny \mbox{on}}^{(t)}$ denotes a support set of $\bchi^{(t+1)} \triangleq \bPsi \bx^{(t)} + \bGamma \by$ that is immune from hard-thresholding, and $\Omega_{\tiny \mbox{off}}^{(t)}$ denotes a second support set that is automatically forced to zero.  The remaining elements of $\bchi^{(t+1)}$ not in $\Omega_{\tiny \mbox{on}}^{(t)} \bigcup \Omega_{\tiny \mbox{off}}^{(t)}$ then face the standard hard-thresholding operator, with all but the largest $k^{(t)}$ values set to zero.  In spirit, (\ref{eq:general_IHT_layer2b}) can be viewed as something like a highway network element \citep{Srivastava15} or a LSTM cell \citep{hochreiter1997long}, where elements can be turned on and off via a gating mechanism separate from the activation function.

When combined with layer-wise weights that change from iteration to iteration via a prescribed sequence (just like a DNN), we arrive at what we term \emph{adaptive} IHT:
\begin{definition}[Adaptive Iterative Hard Thresholding (A-IHT)]
Let $\bx^{(0)} = {\bf 0}$ and assume we have access to some predefined sequence of weights $\{\bPsi^{(t)}, \bGamma^{(t)} \}$ as well as a predefined schedule for computing $\Omega_{\tiny \mbox{on}}^{(t)}$, $\Omega_{\tiny \mbox{off}}^{(t)}$, and $k^{(t)}$.  Then we refer to the iterations
\begin{equation} \label{eq:general_IHT_layer3}
\bx^{(t+1)} = H_{k^{(t)}}\left[\bPsi^{(t)} \bx^{(t)} + \bGamma^{(t)} \by ; \Omega_{\tiny \mbox{on}}^{(t)},\Omega_{\tiny \mbox{off}}^{(t)}  \right]
\end{equation}
as adaptive iterative hard-thresholding.
\end{definition}

We will now examine how A-IHT can directly handle the recovery of maximally sparse signals arising from the clustered dictionary model.  We first introduce some additional notation.  Let $\mathcal{J}$ denote any subset of $\{1,\ldots, c\}$ such that  $ \bA (\mathcal{J}) \triangleq \left[ \bA_j : j \in  \mathcal{J},  \right]$; in other words $\bA(\mathcal{J})$ represents the matrix formed by concatenating all partitions of $\bA$ from the set $\mathcal{J}$.

\begin{proposition} \label{prop:existence1}
Suppose $\bPhi$ is generated from the clustered dictionary model and that the concatenated matrix $[\bU, \bA(\mathcal{J}) ]$ has RIP constant $\delta_{(3k_x + k_c)}\left([\bU, \bA (\mathcal{J}) ]\right) < 1/\sqrt{32}$ for all possible $\mathcal{J}$ with $|\mathcal{J}| \leq c$. Then there exists an A-IHT algorithm  that is guaranteed to produce the correct support pattern of any $\bx^*$ in a finite number of iterations provided that $\by = \bPhi \bx^*$, $\|\bx^*\|_0 \leq k_x$, $|\mathcal{S}_c(\bx^*)| \leq k_c$, and $\epsilon \in (0,\epsilon']$ where $\epsilon'$ is suitably small.

\end{proposition}

The technical nature of Proposition \ref{prop:existence1} belies the simplicity of the actual underlying core idea.  We unpack this result via a few important intuitions:
\begin{itemize}

\item $\bx^*$ itself is also trivially obtained once the support is correctly estimated.  So practically speaking this result guarantees we can recover $\bx^*$ in a finite number of iterations.

\item The integrated dictionary $\bPhi$  can have an arbitrarily large RIP constant as $\epsilon$ grows small such that IHT (or $\ell_1$ minimization, etc.) will likely fail to ever find the correct support.  In fact, it can be proven that IHT will fail with minimal assumptions.\footnote{Technically speaking, the RIP conditions are sufficient but not necessary conditions for success.  Therefore, just because the constant is too high alone does not always guarantee failure.}

    \item In contrast, the sufficient condition for A-IHT to work only depends on coherence involving $\bU$ and $\bA$, the \emph{components} of the clustered dictionary model, not the integrated dictionary $\bPhi$.  In particular, we require that at both the intra-cluster and between-cluster scales, groups of dictionary columns must be reasonably incoherent.

\item  We can simplify the stated conditions by noting that RIP constants can only go up whenever we increase the number of nonzeros or pad a dictionary with extra columns.  Therefore, since $k_x \geq k_c$ and $[\bU, \bA]$ is a superset of the columns from any $[\bU, \bA (\mathcal{J}) ]$, Proposition \ref{prop:existence1} will also hold under the more stringent but easier to digest constraint $\delta_{4k_x}\left([\bU, \bA ]\right) < 1/\sqrt{32}$.  So we pay a small price in the sparsity level multiplier (from $3k$ for regular IHT to $4k$ for A-IHT), but this is offset by the huge gain in working with $[\bU, \bA]$ as opposed to $\bPhi$ as the argument.  And of course in reality we only need this to hold for all of the much smaller dictionary subsets as stipulated in the proposition, a significantly lower bar.

    \item See the proof for details of how the layer weights $\bPsi^{(t)}$ and $\bGamma^{(t)}$, and support sets $\Omega_{\tiny \mbox{on}}^{(t)}$ and $\Omega_{\tiny \mbox{off}}^{(t)}$ can be constructed.  But the core principle is that earlier layers must be tasked with exposing the correct support at the cluster level, without concern for accuracy within each cluster.  Once the correct cluster support has been obtained, later layers can then be charged with estimating the fine-grain details of within-cluster support.  We believe this type of multi-resolution sparse estimation is essential when dealing with highly coherent dictionaries (more on this in the next section).

   \item The support sets $\Omega_{\tiny \mbox{on}}^{(t)}$ and $\Omega_{\tiny \mbox{off}}^{(t)}$ allow the network to `remember' previously learned cluster-level sparsity patterns, in much the same that LSTM gates allow long term dependencies to propagate \citep{hochreiter1997long} or highway networks \citep{Srivastava15} facilitate information flow unfettered to deeper layers.  Moreover, practically speaking these sets can be computed by passing the prior layer's activations $\bx^{(t)}$ through linear filters followed by indicator functions, again reminiscent of how DNN gating functions are typically implemented.

\item Even if we exclude the column normalization multiplier $\bN$ from the clustered dictionary model,  IHT will still fail since the top $k_x$ elements obtained via hard-thresholding will be dominated by columns of $\bPhi$ with large norms at the mercy of $\bv_j$ scale factors.

\end{itemize}

We next turn to the more practically relevant situation where the dictionary cannot be so neatly partitioned into two levels of detail, such that manual construction of a priori layer-dependent weights is much more difficult.

\section{Discriminative Multi-Resolution Sparse Estimation} \label{sec:multi_resolution}

As implied previously, guaranteed success for most existing sparse estimation strategies hinges on the dictionary $\bPhi$ having columns drawn (approximately) from a uniform distribution on the surface of a unit hypersphere, or some similar condition to ensure that subsets of columns behave approximately like an orthogonal basis.  Essentially this confines the structure of the dictionary to operate on a single universal scale.  The clustered dictionary model described in the previous section considers a dictionary built on two different scales, with a cluster-level distribution (coarse) and tightly-packed within-cluster details (fine).  But in reality practical dictionaries may display structure operating across a variety of scales that interleave with one another, forming a continuum among multiple levels.

When the scales are clearly demarcated, we have seen that it is possible to manually define a multi-resolution A-IHT algorithm that guarantees success in recovering the optimal support pattern; and indeed, A-IHT could be extended to handle a clustered dictionary model with nested structures across more than two scales.  However, without clearly partitioned scales it is much less obvious how one would devise an optimal IHT modification.  It is in this context that learning with DNNs is likely to be most advantageous.  In fact, the situation is not at all unlike many computer vision scenarios whereby handcrafted features such as SIFT may work optimally in confined, idealized domains, while learned CNN-based features are often more effective otherwise.

Given a sufficient corpus of $\{\bx^*, \by\}$ pairs linked via some fixed $\bPhi$, we can replace manual filter construction with a learning-based approach.  On this point, although we view our results from Section \ref{sec:adaptive_weights} as a convincing proof of concept, it is unlikely that there is anything intrinsically special about the specific hard-threshold operator and layer-wise construction we employed per se, as long as we alow for deep, adaptable layers that can account for structure at multiple scales.  In practice, we expect that it is more important to establish a robust training pipeline that avoids stalling at the hand of vanishing gradients with deep network structure.  It is here that we propose three significant deviations from the original IHT template.

\begin{enumerate}
\item We exploit the fact that in producing a maximally sparse vector $\bx^*$, the main challenge is estimating $\mbox{supp}[\bx^*]$.  Once the support is obtained, computing the actual nonzero coefficients just boils down to solving something like a least squares problem.  But any learning system will be unaware of this and could easily expend undue effort in attempting to match coefficient magnitudes at the expense of support recovery.  Certainly the use of a data fit penalty of the form $h\left( \| \by - \bPhi \bx \|_2 \right)$, as is adopted by nearly all sparse recovery algorithms, will expose us to this issue.  Therefore we instead formulate sparse recovery as a multi-label classification problem.  More specifically, instead of directly estimating $\bx^*$, we attempt to learn $\bs^* = [s^*_1,\ldots,s^*_m]^{\top}$, where
\begin{equation} \label{eq:s_definition}
s^*_i = 1  ~~ \mbox{if} ~~ x_i^* \neq 0, ~~~ \mbox{and} ~~ s^*_i = 0 ~~ \mbox{otherwise}.
\end{equation}
For this purpose we may then incorporate a traditional multi-label classification loss function via a final softmax output layer, which forces the network to only concern itself with learning support patterns.  This substitution is further justified by the fact that even with traditional IHT, the support pattern will be accurately recovered before the iterations converge exactly to $\bx^*$.  Therefore we may expect that fewer layers (as well as training data) are required if all we seek is a support estimate.

\item Given that IHT can take many iterations to converge on challenging problems, we may expect that a relatively deep network structure will be needed to obtain exact support recovery.  We must therefore take care to avoid premature convergence to areas with vanishing gradient by incorporating several recent countermeasures proposed in the DNN community.  For example, as mentioned previously, the adaptive threshold operator A-IHT employs is reminiscent of highway networks or LSTM cells, which have been proposed to allow longer range flow of gradient information to improve convergence through the use of gating functions.  An even simpler version of this concept involves direct, un-gated connections that allow much deeper `residual' networks to be trained \citep{he2015deep} (which is also reminiscent of the residual factor embedded in the original IHT iterations). We deploy this tool, along with batch-normalization \citep{ioffe2015batch} to aid convergence, for our basic feedforward pipeline.  Later we also consider an alternative structure based on recurrent LSTM cells.  Note that unfolded LSTM networks frequently receive a novel input for every time step, whereas here $\by$ is applied unaltered at every layer (more on this in Section \ref{sec:lstm}).

    \item We replace the non-integrable hard-threshold operator with simple rectilinear (ReLu) units \citep{Nair10}, which are functionally equivalent to one-sided soft-thresholding.
\end{enumerate}

Taken together, these changes deviate from the original IHT script; however, we believe they nonetheless preserve the foundational principles of learning-based sparse estimation.  Certainly our empirical evidence below supports this claim.

\section{Construction of Training Sets} \label{sec:training}
If the ultimate goal is to learn an accurate model for computing the minimum $\ell_0$-norm, maximally sparse solution, then care must be taken in how we construct the training data.  This issue is especially acute in the operational regime considered herein, namely sparse linear inverse problems where dictionary coherence is high.

As an illustrative example of this point, suppose we have a dictionary $\bPhi$ that is known to facilitate highly compact representations of some signal class of interest $\mathcal{Y}$.  Even if we have access to a large set of observations $\by \in \mathcal{Y}$, our work is still ahead of us to construct a viable training set.  This is because in general, computing the maximally sparse $\bx^*$ that corresponds with each $\by$ represents an NP-hard problem, and if the dictionary is coherent fast approximate schemes like OMP and IHT will fail; similarly $\ell_1$-$\ell_0$ norm equivalency breaks down corrupting convex solutions.  Consequently, if we use any of these methods to generate training values for $\bx^*$, we will merely end up learning a network that approximates a suboptimal strategy, not the true $\ell_0$ norm solution that represents our goal to begin with. To the best of our knowledge though, this is the route that all previous learning-based sparse estimation pipelines proceed, e.g., see \citep{Gregor10,Sprechmann15,Wang15l0}.  Hence these systems do not actually produce maximally sparse estimates as is our focus here, although they do represent quite useful methods for reducing the computational burden of approximate schemes like $\ell_1$ minimization.

In this paper we advocate an entirely different strategy.  We first generate sparse training vectors $\bx^*$ with random support patterns.  We then compute synthetic observations using $\by = \bPhi \bx^*$ (possibly with additional additive noise for robustness).  Provided the dictionary satisfies minimal assumptions related to a quantity called matrix spark \citep{Donoho03}, $\bx^*$ will provably represent the maximally sparse feasible solution.  In this way we have an inexpensive means of producing whatever volume of training data we desire.  Moreover, we have observed modest sensitivity at test time to the actual magnitude distribution of nonzero coefficients in $\bx^*$ used during training.  In other words, even if we test using a different magnitude distribution than was used to generate training data, performance is relatively stable (see Section \ref{sec:simulations}).  This is a likely consequence of using a softmax final multi-label classification layer as opposed to trying to directly estimate $\bx^*$.  In practice this stability is paramount since we may not have a good estimate of this distribution anyway.

\section{Feedforward Network Experiments} \label{sec:simulations}


With existing sparse optimization algorithms, the goal is to estimate $\bx^*$ when presented with $\by$ and $\bPhi$.  In contrast, with a large corpus of training pairs $\{\bx^*,\by\}$ we intend to learn a mapping from $\by$ to $\bx^*$ using a deep architecture.  To isolate the various factors affecting the performance of feedforward networks in particular, this section describes experiments using a variety of different data generation procedures.  Later, Section \ref{sec:lstm} will consider a competing recurrent LSTM architecture.

\subsection{Network Design}
\begin{figure}
    \centering
        \includegraphics[width=0.95\columnwidth]{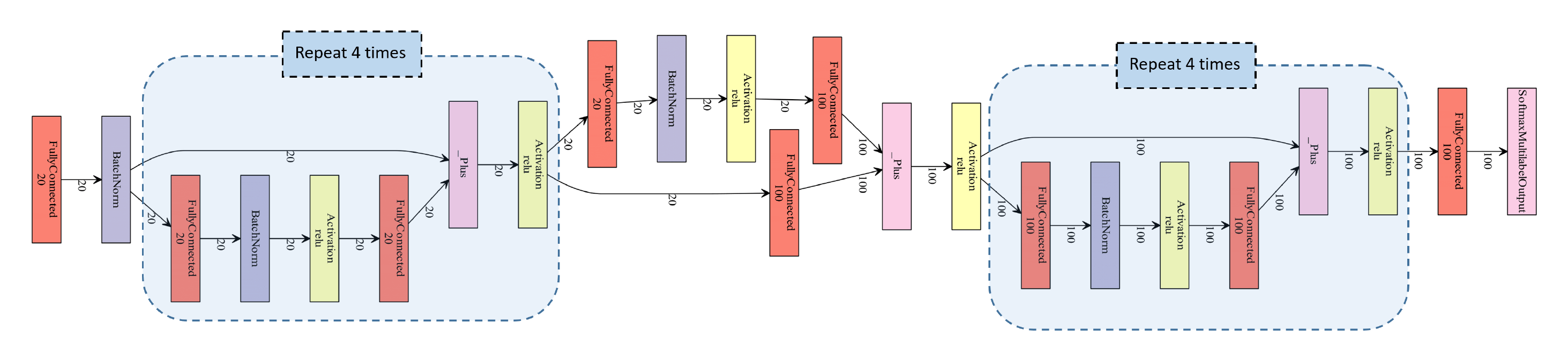}
     \caption{Basic network structure (zoom to view details).}
\label{fig:netstr}
\end{figure}

We confine our design here to the feedforward structure motivated in Section \ref{sec:multi_resolution}.  In brief, we build a 20-layer network with residual connections \citep{he2015deep} and batch normalization \citep{ioffe2015batch}. Moreover, because our sparse data will ultimately have no indication of local smoothness, we use fully connected layers rather than convolutions.  For the nonlinearities we apply rectilinear units (ReLU).  We also include a final softmax layer which outputs a vector $\bp \in \mathbb{R}^{m}$, with $p_j \in [0,1]$ providing an estimate of the probability that $x_j \neq 0$. The detailed network structure, which was implemented using MXNet \citep{chen2015mxnet}, can be found in Figure \ref{fig:netstr}.

\subsection{Basic Sparse Estimation Experimental Setup}
\label{ssec:basic}

We generate a dictionary matrix $\bPhi \in \mathbb{R}^{n\times m}$ using
\begin{equation}
\bPhi =\sum_{i=1}^{n} { \frac{1}{i^2}  \bu  \bv^{\top }},
\end{equation}
where $\bu \in \mathbb{R}^n$ and $\bv \in \mathbb{R}^m$ have iid elements drawn from $\mathcal{N}(0,1)$.  We also rescale each column of $\bPhi$ to have unit $\ell_2$ norm. $\bPhi$ generated in this way has super-linear decaying singular values (indicating correlation between the columns) but is not constrained to any specific structure. Many dictionaries in real applications have such a property. As a basic experiment, we generate $N$ ground truth samples $\bx^* \in \mathbb{R}^m$ by randomly selecting $d$ nonzero entries, with nonzero amplitudes drawn iid from the uniform distribution $\mathcal{U}[-0.5,0.5]$, excluding the interval $[-0.1,0.1]$ to avoid small, relatively inconsequential contributions to the support pattern.  We then create $\by \in \mathbb{R}^n$ via $\by = \bPhi \bx^*$. As we increase $d$, the sparse estimation problem becomes intrinsically more difficult.  We set $n=20$ and $m=100$, while $d \le 10$, noting that if $d=10$ we have only twice as many measurements as nonzeros in $\bx^*$, which is a challenging regime, especially when $\bPhi$ has strong correlations.

We generated $N=700000$ total samples and used the first $N_1=600000$ for training and the remaining $N_2=100000$  for testing. Network optimization is achieved via stochastic gradient descent (SGD) with a momentum of 0.9 and weight decay of 0.0001. The initialization follows \citep{he2015delving} and we did not apply any drop-out. The batch size was set to 250. The initial learning rate was 0.01 and was reduced by 90\% every 50 epoches. We stopped training after 150 epoches, at which point empirical convergence was always observed.  Unless otherwise specified, throughout this paper we convert $\bx^*$ to the binary label vector $\bs^*$ from (\ref{eq:s_definition}) for training purposes using the stated softmax output layer.

To evaluate the performance, we introduce two metrics referred to as \emph{strict accuracy} (s-acc) and \emph{loose accuracy} (l-acc), respectively. Both depend on two sets for each sample/trial, the ground truth labels and the predicted top-$d$ labels given by
\begin{eqnarray}\label{eq:sets}
\mathcal{S}_{gt} & = & \{j:x_j \neq 0 \} \nonumber \\
\mathcal{S}_{pred}(d) & = & \{j: p_j \mbox{ is one of the } d \mbox{ largest values.}\}
\end{eqnarray}
Strict accuracy evaluates whether the $d$ ground truth nonzeros are exactly aligned with the predicted top-$d$ values produced by our network, and when averaged across test trails, can be computed via
\begin{equation}\label{eq:s-acc}
    \text{s-acc}=\frac{1}{N_2}\sum_{i=1}^{N_2} {  \mathbb{I}\left[ \mathcal{S}_{gt}^{(i)} = \mathcal{S}_{pred}^{(i)}(d) \right]  }
\end{equation}
where $\mathbb{I}[\cdot]$ is an indicator function and here the superscript $(i)$ denotes the sample number.  This all-or-nothing metric is commonly adopted in the compressive literature, and effectively measures the percentage of trials where we can perfectly recover $\bx^*$.

In contrast, loose accuracy considers the degree to which the correct support is included in the largest values of $\bp$.  Note that given our experimental design, it can be shown that with probability one there exists only a single feasible solution to $\by = \bPhi \bx$ such that $\| \bx \|_0 < n = 20$ (this will define the optimal support set by design).  Moreover, any support pattern with exactly $n$ nonzeros is sufficient to produce a unique feasible solution (referred to as a basic feasible solution in the linear programming literature \citep{Luenberger84}).  Hence we define loose accuracy as the degree to which the true support indeces are contained within the top 20 largest values of $\bp$, or
\begin{equation}\label{eq:l-acc}
    \text{l-acc}=\frac{1}{N_2}\sum_{i=1}^{N_2} \frac{ \left|\mathcal{S}_{gt}^{(i)} \cup \mathcal{S}_{pred}^{(i)}(n) \right| }{ d }.
\end{equation}

Both s-acc and l-acc metrics are computed for all $N_2$ test points and compared against a battery of existing algorithms, both learning- and optimization-based.\footnote{For competing algorithms, we compute $\mathcal{S}_{pred}(d)$ using the largest (in magnitude) $n$ elements of any estimate $\hat{\bx}$. }  These include standard $\ell_1$ minimization via ISTA iterations \citep{Candes06}, IHT \citep{Blumensath09}, an ISTA-based network \citep{Gregor10}, and an IHT-inspired network \citep{Wang15l0}.  For $\ell_1$ minimization we used publicly-available ISTA code,\footnote{http://www.eecs.berkeley.edu/~yang/software/l1benchmark/index.html} while for IHT we applied our own implementation (it only requires a few lines in Matlab) and supplied the hard-thresholding operator with the ground truth number of nonzeros, meaning $k = d$ for all experiments.  For the learning-based methods, we estimated parameters for both the ISTA- and IHT-based networks using MXNet and the exact same training data described above.

\begin{figure}
    \centering
        \includegraphics[width=0.95\columnwidth]{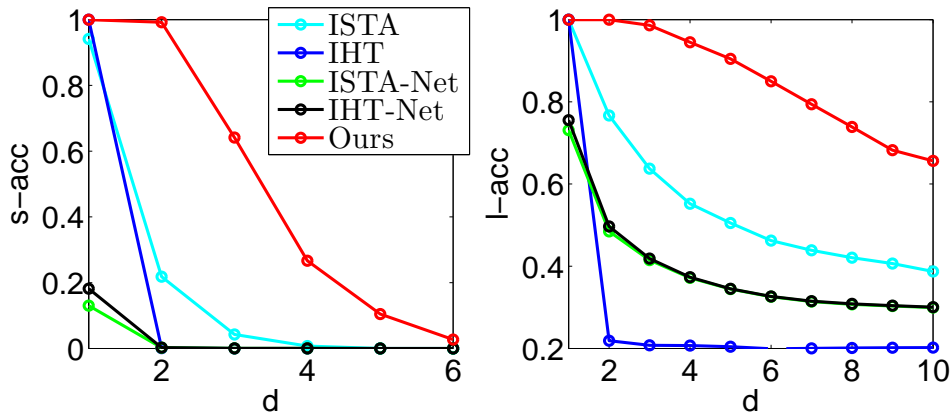}
     \caption{Support recovery accuracy with uniformly distributed nonzero elements. \emph{Left}: strict accuracy, \emph{Right}: loose accuracy. Note that under the stated design conditions with $m = 5n$, random guessing will lead to a loose accuracy value of 0.20.}
\label{fig:eva1}
\end{figure}

\subsubsection{Accuracy Results}

Figure \ref{fig:eva1} illustrates how different methods perform under both evaluation metrics.  Given the correlated $\bPhi$ matrix, the recovery performance of IHT, and to a lesser degree $\ell_l$ minimization using ISTA, is rather modest as expected given that the associated RIP constant will be quite large by construction.  In contrast our method achieves uniformly higher accuracy under both metrics, including over existing learning-based methods trained with the same data.  This improvement is likely the result of three significant factors: (i) Existing learning methods initialize using weights derived from the original sparse estimation algorithms, but such an initialization will be associated with locally optimal solutions in most cases with correlated dictionaries. (ii) As described in Section \ref{sec:constant_weights}, constant weights across layers have limited capacity to unravel multi-resolution dictionary structure, especially one that is not confined to only possess some low rank correlating component. (iii) The quadratic loss function used by existing methods does not adequately focus resources on the crux of the problem, which is accurate support recovery.  In contrast our approach adopts an initialization motivated directly by DNN-based training considerations, unique layer weights to handle a multi-resolution dictionary, and a multi-label classification output layer to focus learning on support recovery.

\subsubsection{Computational Efficiency}

Table \ref{tab:runtime} displays the average per-sample runtime required to produce each sparse estimate.  Not surprisingly, the learning-based methods display a dramatic advantage over ISTA-based $\ell_1$-minimization and IHT, both of which require a high number of iterations to converge.  In contrast, ISTA-Net, IHT-Net, and our method only involve passing activations through a handful of layers.  Although these learning-based approaches all require a potentially expensive training phase, for any task of interest with a fixed $\bPhi$ matrix, we need only fit the network model once up front and then subsequent testing/deployment will always be much more efficient.

\begin{table}
\caption{Average per-sample runtime to produce sparse estimates (in seconds).}
\vspace*{0.2cm}
\label{tab:runtime}
\begin{center}
\begin{tabular}{c||M{17mm}|M{17mm}|M{17mm}|M{17mm}|M{17mm}}
\hline
  & ISTA & IHT & ISTA-Net & IHT-Net & Ours  \\
\hline
runtime  & 0.7393 & 0.288 & 8.17$\times 10^{-7}$ & 8.68$\times 10^{-7}$ & 9.71$\times 10^{-7}$ \\
\hline
\end{tabular}
\end{center}
\end{table}

\begin{figure}
    \centering
        \includegraphics[width=0.95\columnwidth]{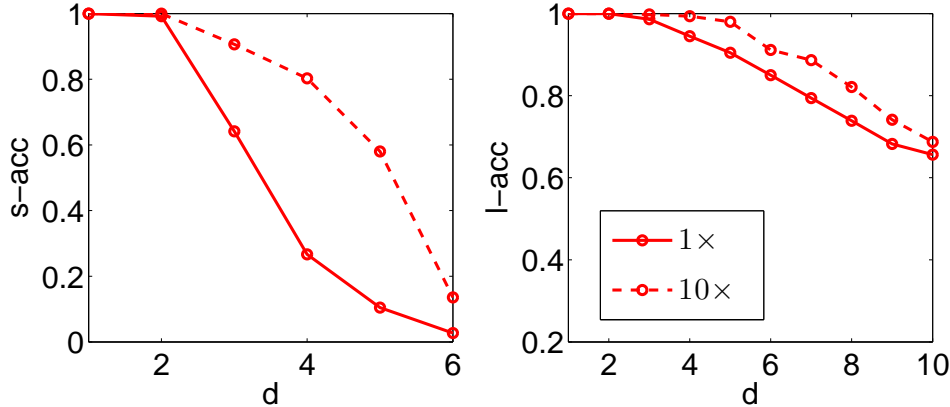}
     \caption{Accuracy results with $1\times$ and $10\times$ the original number of training data.}
\label{fig:case1}
\end{figure}

\subsection{Variants of the Basic Experiment}
\label{ssec:real}

Here we vary a number of different factors from the basic experiment, in each case holding all others fixed.

\subsubsection{Varying training set size}

In Figure \ref{fig:case1}, we see that adding more training data to our method can further boost accuracy. This is not a surprise given that it is fundamentally based on learning, and therefore, as long as the capacity of the network allows and optimization ends up in a good basin, we may expect some improvement with additional data.

\subsubsection{Alternative network structures}

\begin{figure}
    \centering
        \includegraphics[width=0.95\columnwidth]{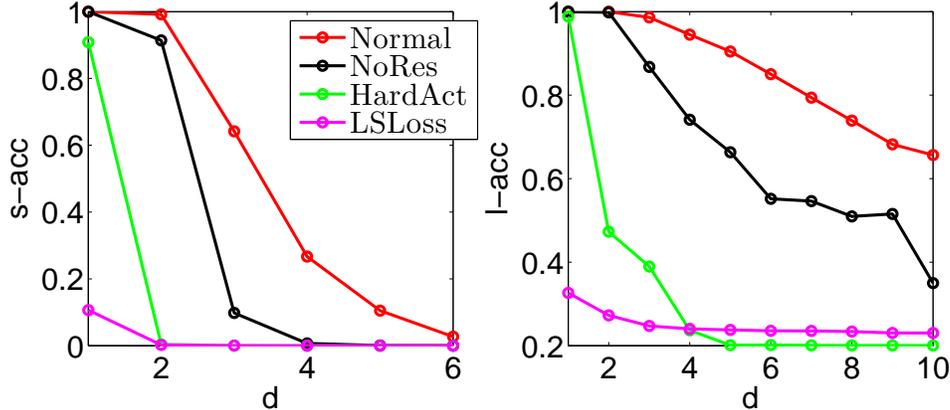}
     \caption{Comparison of our baseline network with an equivalent network using (i) no residual connections, (ii) hard-threshold activation function, and (iii) least-squares loss.  Note that each change is applied in isolation, not in aggregate.}
\label{fig:case234}
\end{figure}

As discussed in Section \ref{sec:multi_resolution}, our DNN design choices were largely motivated by practical issues related to the information and gradient flows to which DNN training can be highly sensitive. In this section we examine different network architectures to quantify essential factors affecting performance.  In particular, we consider the following changes:
\begin{enumerate}
\item We remove the residual-net connections.
\item We replace ReLU with hard-threshold activations. In particular, we utilize the so-called HELU$_\sigma$ function introduced in \citep{Wang15l0}, which is a continuous and piecewise linear approximation of the scalar hard-threshold operator given by
    \begin{equation*}
      HELU_{\sigma}(x)=\begin{cases}
               0~~~~~~~~~~~~~~~~~~~~\mbox{if}~~|x|\le1-\sigma\\
               \frac{1}{\sigma}(x-1+\sigma) ~~~~~\mbox{if}~~1-\sigma < x <1\\
               \frac{1}{\sigma}(x+1-\sigma) ~~~~~\mbox{if}~~-1< x <\sigma-1\\
               \bu_i ~~~~~~~~~~~~~~~~~~~\mbox{if}~~|x|\ge1.
            \end{cases}
    \end{equation*}
\item We use a quadratic penalty layer instead of a multi-label classification loss layer, i.e., the loss function is changed to $\sum_{i=1}^{N_1}\| \ba^{(i)}-\by^{(i)} \|_2^2$ (where $\ba$ is the output of the last fully-connected layer) during training.
\end{enumerate}
Figure \ref{fig:case234} displays the associated recovery percentages, where we observe that in each case performance degrades.  Without the residual design, and also with the inclusion of a rigid, non-convex hard-threshold operator, local minima during training appear to be a likely culprit, consistent with observations from \citep{he2015deep}.  Likewise, use of a least-squares loss function is likely to emphasize the estimation of coefficient amplitudes rather than focusing on support recovery.

\subsubsection{Different distributions for $\bx^*$}

From a practical standpoint, we would like to estimate the support pattern of all the significant elements of $\bx^*$; however, these elements need not all have the same amplitudes.  Moreover, in practice we may expect that the true amplitude distribution may deviate at times from the original training set.  To explore robustness to such mismatch, as well as different amplitude distributions, we consider two sets of candidate data: the original from Section \ref{ssec:basic}, and similarly-generated data but with the uniform distribution of nonzero elements replaced with the Gaussians $\mathcal{N}(\pm 0.3,0.1)$, where the mean is selected with equal probability as either $-0.3$ or $0.3$, thus avoiding tiny magnitudes with high probability.

\begin{figure}
    \centering
        \includegraphics[width=0.95\columnwidth]{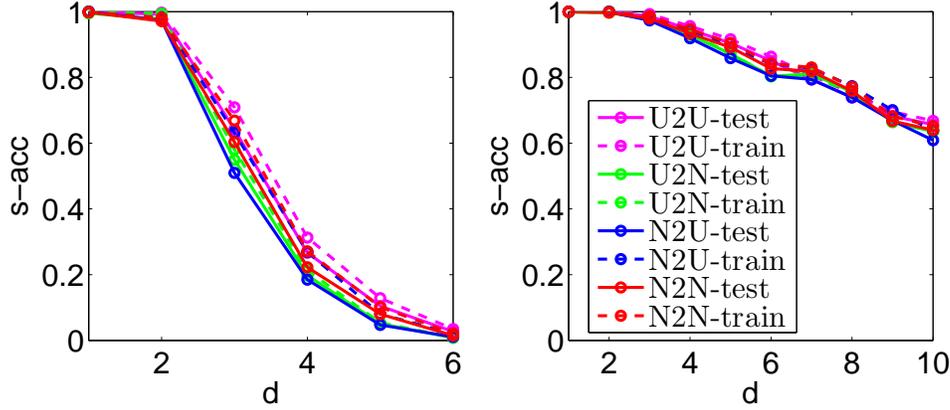}
     \caption{Accuracy under different continuous training and testing distributions.}
\label{fig:case5}
\end{figure}

Figure \ref{fig:case5} reports accuracies under different distributions for both training and testing, including mismatched cases.  The label `U2U' refers to training and testing with the uniformly distributed amplitudes described in Section \ref{ssec:basic}, while `U2N' uses uniform training set and a Gaussian test set.  Analogous definitions apply for `N2N' and `N2U'.  In all cases we note that the performance is quite stable across training and testing conditions.  We would argue that our recasting of the problem as multi-label classification contributes, at least in part, to this robustness. The application example described next demonstrates further tolerance of training-testing set mismatches.

\section{Practical Application: Photometric Stereo} \label{sec:application}


Photometric stereo represents a powerful technique for recovering high-resolution surface normals from a 3D scene using appearance variations in 2D images under different lightings. For example, when images of an ideal Lambertian surface are obtained under illumination from three known directions, the surface orientation can be uniquely determined using a simple least-squares fit \citep{Woodham80}.

In practice however, the estimation process is often disrupted by non-Lambertian effects such as specular highlights, shadows, or image noise. To account for such outlying factors, robust estimation methods have been proposed that decompose an observation matrix of stacked images under different lighting conditions into an ideal Lambertian component and a sparse error term \citep{Wu10,Ikehata12}.  While principled in theory, this approach requires solving on the order of $10^4-10^6$ distinct sparse regression problems, one for each point for which we would like to obtain a surface normal estimate.  We will now map this application domain into our sparse DNN framework, which can readily handle the required outlier removal problem potentially orders-of-magnitude faster than existing practical systems, facilitating real-time deployment in mobile environments.

\subsection{Problem Details}
\label{ssec:psvsp}
Suppose we have $q$ observations of a given surface point from a Lambertian scene under different lighting directions.  Then the resulting measurements, denoted $\bo \in \mathbb{R}^{q}$, can be expressed as
\begin{equation}
\bo = \rho \bL \bn,
\end{equation}
where $\bn \in \mathbb{R}^{3}$ denotes the true surface normal, each row of $\bL  \in \mathbb{R}^{q\times 3}$ defines a lighting direction, and $\rho$ is the diffuse albedo, acting here as a scalar multiplier \citep{Woodham80}.  If specular highlights, shadows, or other gross outliers are present, then the observations are more realistically modeled as
\begin{equation}
\bo = \rho \bL \bn + \bolde,
\end{equation}
where $\bolde$ is an an unknown sparse vector \citep{Wu10,Ikehata12}.  In this revised scenario, we might consider estimating $\bn$ using
\begin{equation}\label{eq:point}
  \min_{\widetilde{\bn},\bolde }  \| \bolde \|_0 ~~~ \mbox{s.t.} ~ \bo = \bL \widetilde{\bn} + \bolde,
\end{equation}
where $\widetilde{\bn}$ is merely the surface normal rescaled with $\rho$.  From this expression, it is apparent that, since $\widetilde{\bn}$ is unconstrained, $\bolde$ need not compensate for any component of $\bo$ in the range of $\bL$.  Given that $\mbox{null}[\bL^{\top}]$ is the orthogonal complement to $\mbox{range}[\bL]$, we may transform (\ref{eq:point}) to the equivalent problem
\begin{equation}\label{eq:point2}
  \min_{\bolde } \| \bolde \|_0 ~~~ \mbox{s.t.} ~ \mbox{Proj}_{\mbox{null}[L^{\top}]} (\bo) = \mbox{Proj}_{\mbox{null}[L^{\top}]} (\bolde).
\end{equation}
The constraint is of course equivalent to $\by = \bPhi \bolde$ with $\by = \mbox{Proj}_{\mbox{null}[L^{\top}]} (\bo)$ and $\bPhi \bolde = \mbox{Proj}_{\mbox{null}[L^{\top}]} (\bolde)$, and so (\ref{eq:point}) ultimately collapses to our canonical sparse estimation problem from (\ref{eq:basic_L0_prob}).  Additionally, given that rows of $\bPhi$ will form a basis for $\mbox{null}[L^{\top}]$ which is lighting-hardware dependent, there are likely to be unavoidable correlations in the dictionary columns.

While existing sparse estimation algorithms can be adopted to solve (\ref{eq:point2}), this is impractical for many real-world applications since the number of surface points can be extremely large (possibly even greater than $10^6$ for high-resolution reconstructions).  Fortunately though, given that $\bPhi$ is fixed across all possible scenes and surface points for a given lighting geometry, to apply our method we only need learn a single DNN model, and once trained, testing on novel scenes will be extremely efficient.  This allows fast computation of outlier positions via the support of $\bolde$, after which the remaining inlier points can be used to compute surface normals using a traditional least squares fit.

\subsection{Results}
\label{ssec:app-expr}

Following \citep{Ikehata12}, we use 32-bit HDR gray-scale images of the object \emph{Bunny} (256$\times$256) with foreground masks under different lighting conditions whose directions, or rows of $\bL$, are randomly selected from a hemisphere with the object placed at the center.  To apply our method, we first compute $\bPhi$ using the appropriate projection operator derived from the lighting matrix $\bL$.  As real-world training data is expensive to acquire, we instead synthetically generate a training set as follows.  First, we draw a support pattern for $\bolde$ uniformly at random with cardinality $d$ sampled uniformly from the range $[d_1,d_2]$.  The values of $d_1$ and $d_2$ can be tuned in practice.  Nonzero values of $\bolde$ are assigned iid random values from a Gaussian distribution whose mean and variance are also tunable.  Beyond this, no attempt was made to match the true outlier distributions encountered in applications of photometric stereo. Finally, for each $\bolde$ we can naturally compute observations $\by = \bPhi \bolde$, which serve as candidate network inputs.

\begin{table}
\caption{Photometric stereo results using different methods.}
\vspace*{0.2cm}
\label{tab:ps}
\begin{center}
\begin{tabular}{M{4mm}||M{8mm}|M{8mm}|M{8mm}|M{8mm}|M{8mm}}
\hline
  & \multicolumn{5}{M{54mm}}{Average angular error (degrees)}\\
\hline
 $q$ & LS & $\ell_1$ & SBL & Rnd4 & Ours \\
\hline
40  & 9.33   & 1.24 & 0.50 & 16.64 & 1.20\\
20  & 10.80  & 3.96 & 1.86 & 18.61 & 1.95 \\
10  & 12.13  & 7.10 & 4.02 & 16.56 & 1.48 \\
\hline
\end{tabular}
\vspace{2mm}

\begin{tabular}{M{4mm}||M{8mm}|M{8mm}|M{8mm}|M{8mm}|M{8mm}}
\hline
  &\multicolumn{5}{M{54mm}}{Runtime (sec.)}\\
\hline
 $q$ & LS & $\ell_1$ & SBL & Rnd4 & Ours \\
\hline
40  &  7.84 & 34.9 & 75.2 & 0.68 & 1.25  \\
20  & 5.47 & 34.3 & 42.5  & 0.66 & 1.21\\
10  &4.10 & 33.7 & 59.1& 0.64 & 1.17 \\
\hline
\end{tabular}
\end{center}
\end{table}

Given synthetic training data acquired in this way, we learn a network with the exact same structure and optimization parameters as in Section \ref{sec:simulations}; no application-specific tuning was introduced.  We then deploy the resulting network on the gray-scale Bunny images.\footnote{Note that for each different number of images we must train a separate model, since the lighting geometry effectively changes.  However, in practice this is not an issue since we only ever need train a single model per hardware configuration.}  For each surface point, we use our DNN model to approximate (\ref{eq:point2}).  Since the network output will be a probability map for the outlier support set instead of the actual values of $\mathbf{e}$, we choose the 4 indices with the \emph{least} probability as inliers and use them to compute $\bn$ via least squares.

\begin{figure}
    \centering
    \begin{minipage}{0.33\textwidth}
    \centering
    \subfigure[GT]{
        \includegraphics[width=1\columnwidth]{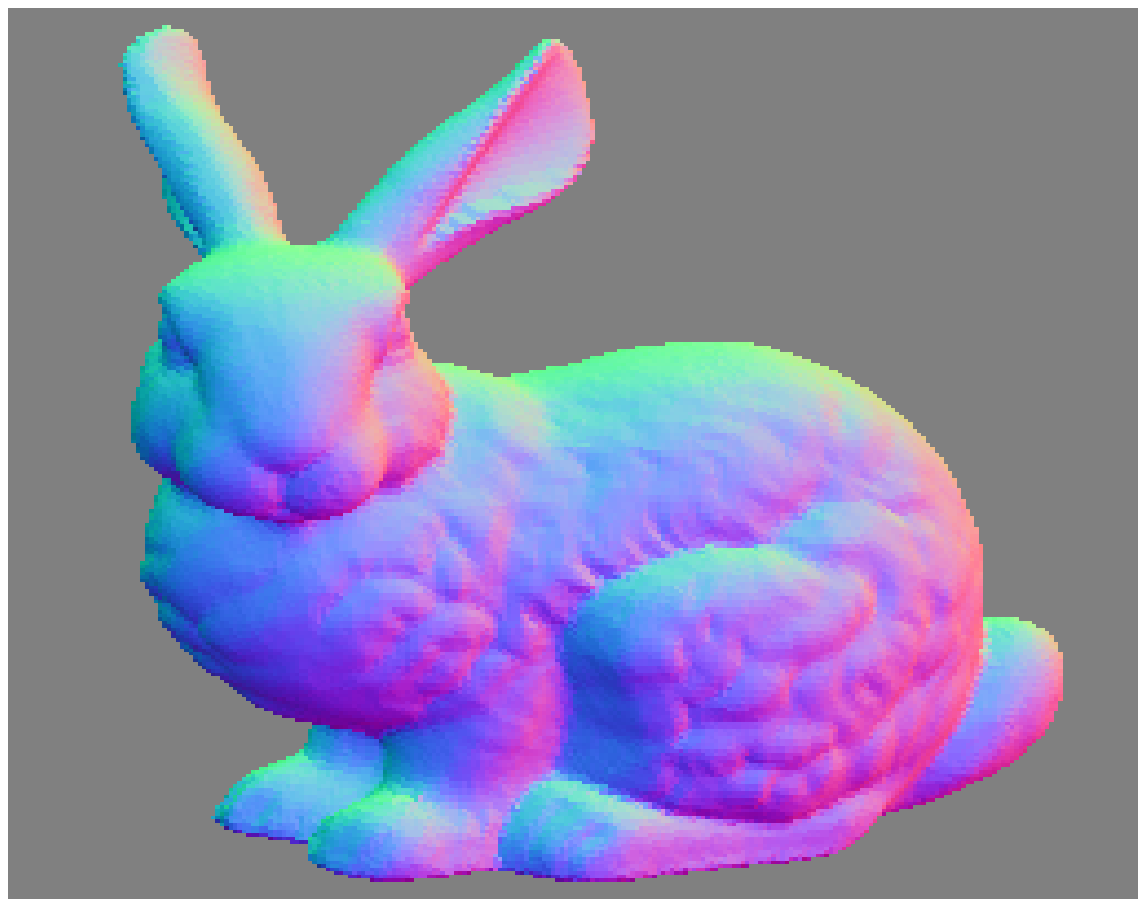}
        }
    \end{minipage}
    \begin{minipage}{0.66\textwidth}
    \centering
    {
        {
    \subfigure[LS]{
        \includegraphics[width=.49\columnwidth]{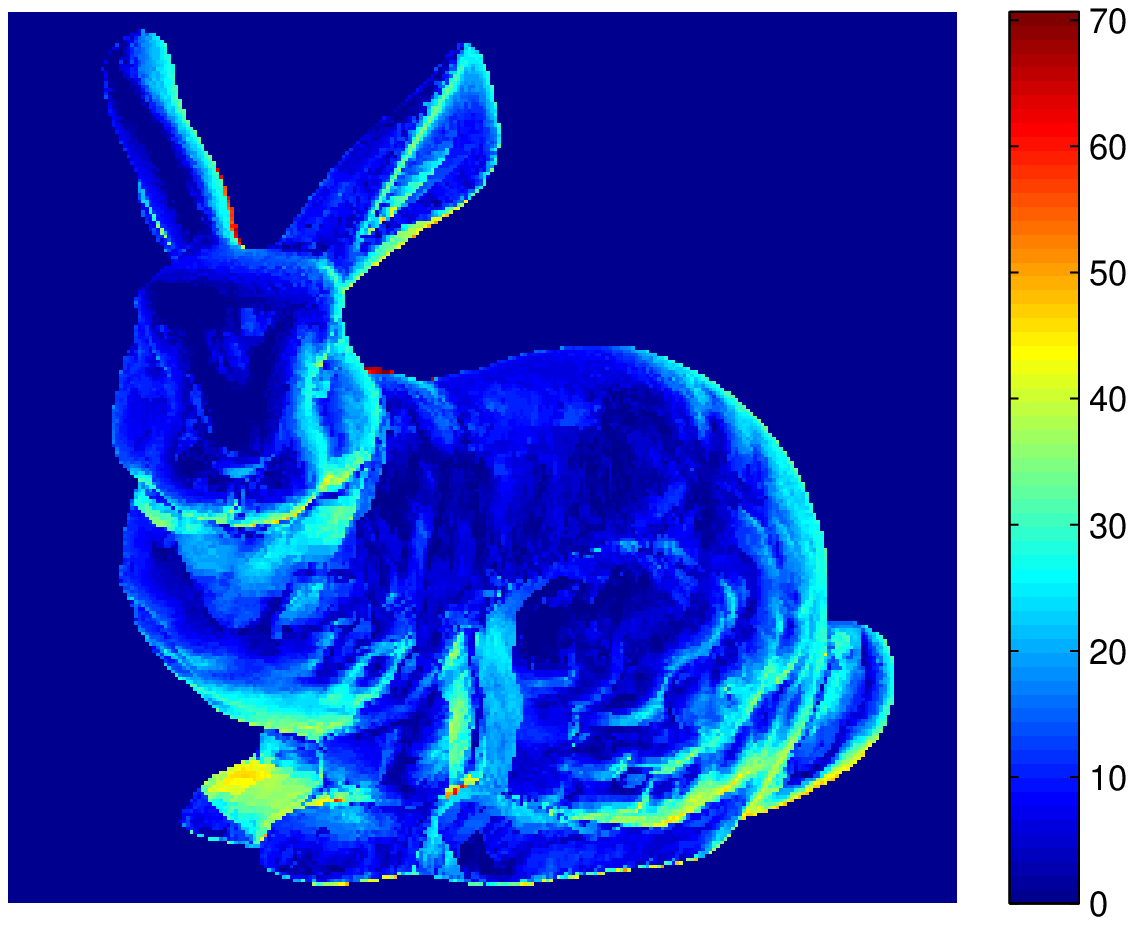}
        }
    \hspace{-7mm}
    \subfigure[$\ell_1$]{
        \includegraphics[width=.49\columnwidth]{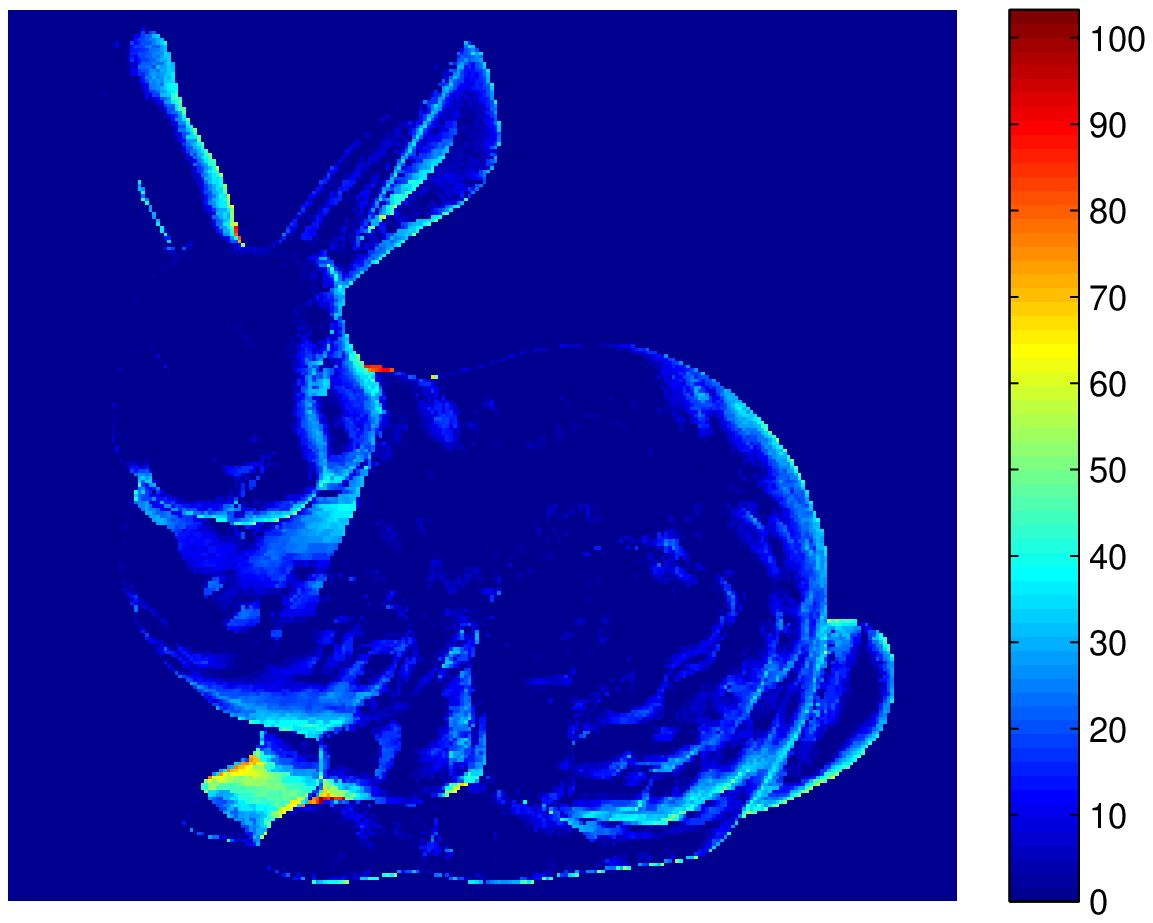}
        }
        }
        {
    \subfigure[SBL]{
        \includegraphics[width=.49\columnwidth]{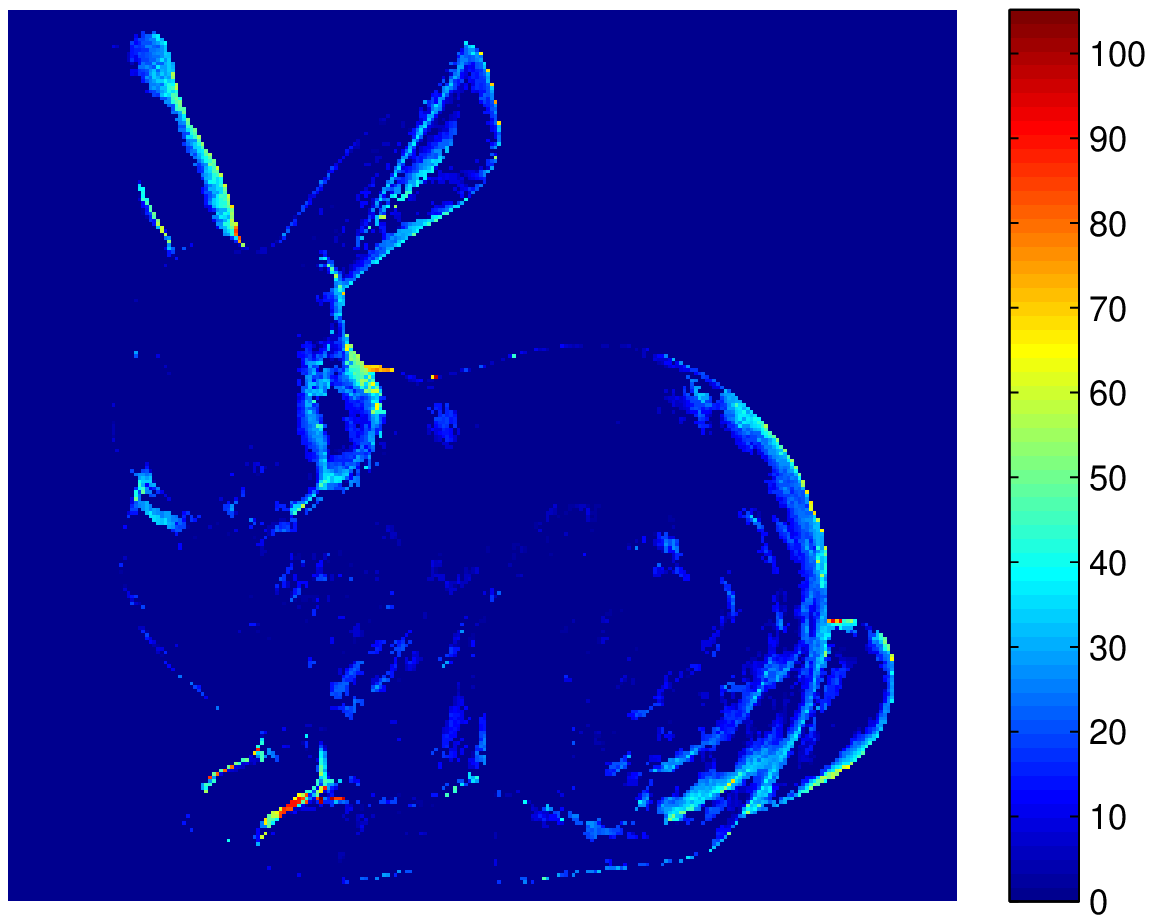}
        }
    \hspace{-7mm}
    \subfigure[Ours]{
        \includegraphics[width=.49\columnwidth]{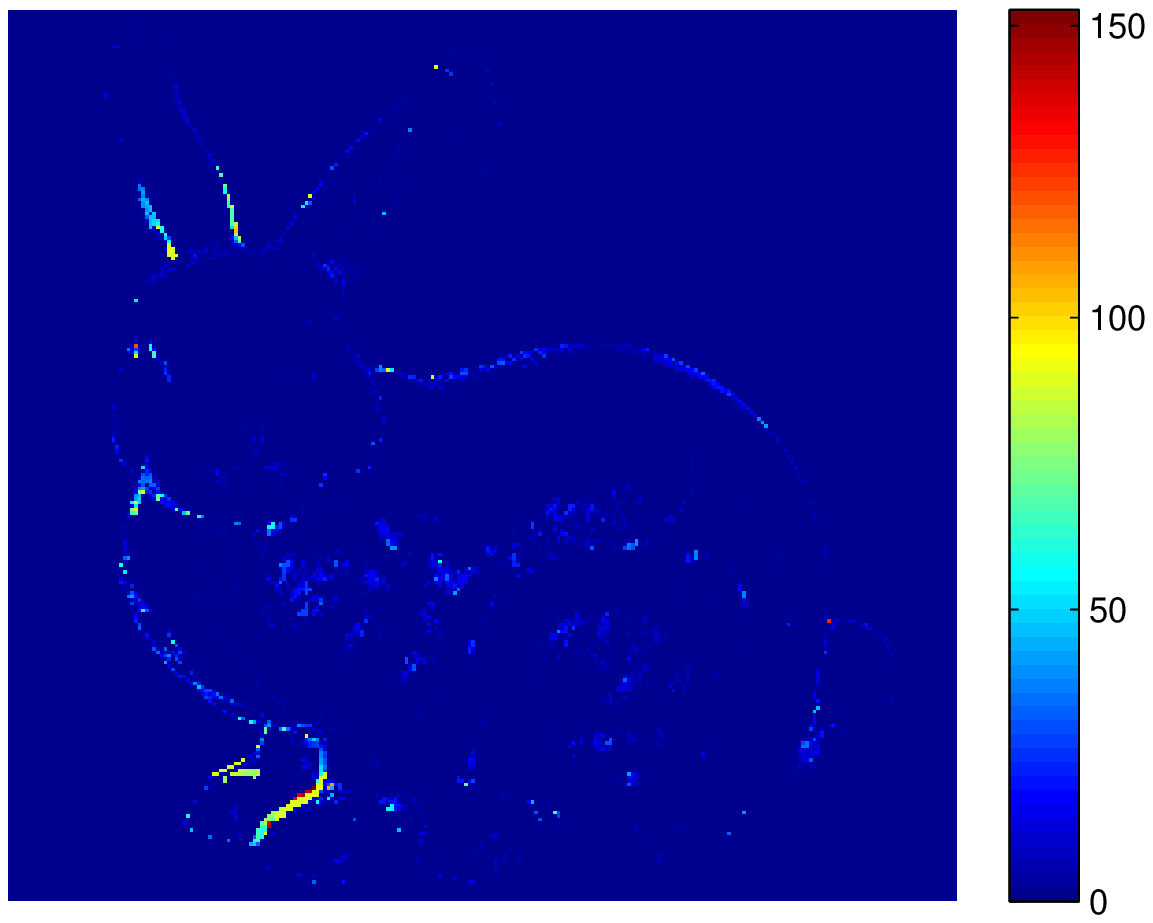}
        }
        }
    }
    \end{minipage}
     \caption{\small{Reconstruction error maps.}}
\label{fig:ps}
\end{figure}

We compare our method against the baseline least squares estimate from \citep{Woodham80}, $\ell_1$ norm minimization, and a sparse Bayesian learning (SBL) approach specifically developed in \citep{Ikehata12} for surface normal estimation.  We also consider a second baseline estimator, denoted `Rnd4', which computes a surface normal estimate using 4 randomly selected indices as putative inliers.  As the number of images $q$ is varied, we compute the angular error between the recovered normal map by each algorithm and the ground truth.

Results are reported in Table \ref{tab:ps} for $q \in \{10,20,40\}$, which also includes runtime comparisons.  In the hardest case, where only $q = 10$ images are present, our method significantly outperforms the others.  For $q \in \{20,40\}$ images our method is still quite competitive, with only SBL offering superior results.  However, it must be noted that SBL represents a  computationally-expensive Bayesian approach especially designed for this problem, with runtimes nearly two orders of magnitude higher than our DNN.\footnote{In fact, the runtime of our method is only twice that of Rnd4, which uses just a single, low-dimensional least squares fit.}  Additionally, a key reason that our approach does not improve substantially as $q$ increases is that we fixed the assumed number of inliers to be 4 in all cases; however, allowing a flexible number that grows with the number of images (as implicitly permitted by SBL) will likely improve performance.

As a complementary perspective, recovered surface normal error maps are displayed in Figure \ref{fig:ps} when $q = 10$.  Here we observe that our DNN estimates lead to far fewer regions of significant error.  Overall though, this application example illustrates that mismatched synthetic training data can, at least for some problem domains, be sufficient to learn a quite useful sparse estimation DNN.

\section{Alternative LSTM Networks} \label{sec:lstm}


Thus far our experimentation has focused on feedforward networks with a residual design.  In this section we turn to a recurrent LSTM structure and execute some preliminary evaluations.  As high-level motivation, there are many similarities between unfolded sparse estimation algorithms like the adaptive IHT discussed in Section \ref{sec:adaptive_weights} and an unfolded LSTM network.  Both supply an input $\by$ to every unfolded layer, and both implicitly utilize gating functions to switch activations on or off as learning proceeds, allowing partial support patterns or other information to be remembered in deeper layers.  Although we will defer a much more detailed, self-contained exploration to a future paper, we nonetheless here present an initial empirical proof-of-concept using a vanilla form of LSTM network that has not been explicitly tailored for sparse estimation problems beyond the  final multi-label classification layer described previously.

\begin{figure}[h]
    \centering
    {
        \subfigure[LSTM-Net]{
        \includegraphics[width=1\columnwidth]{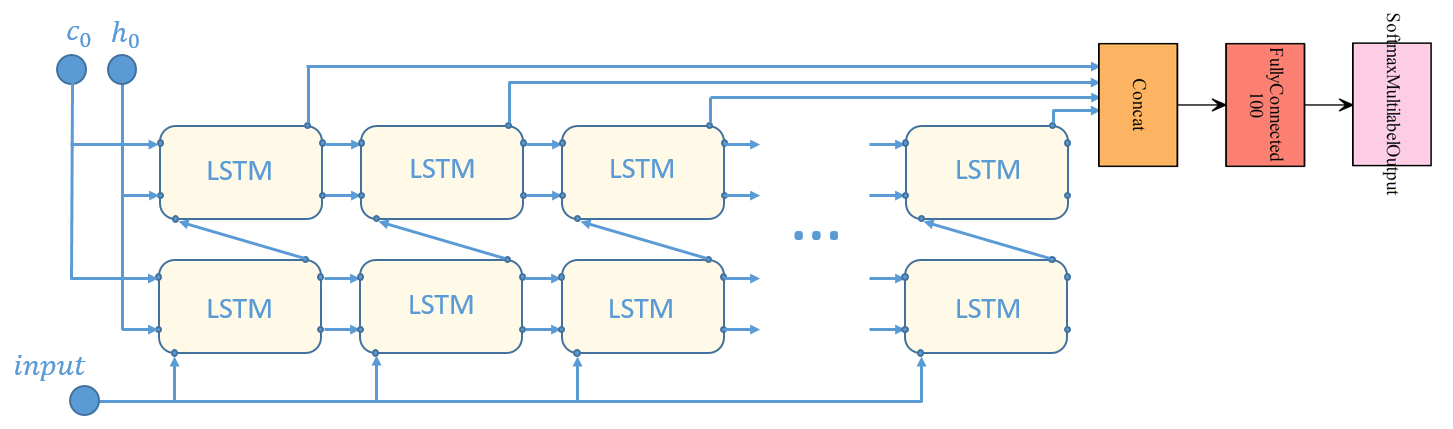}
        }
    \vspace{-2mm}
    \subfigure[LSTM-Cell]{
        \includegraphics[width=0.7\columnwidth]{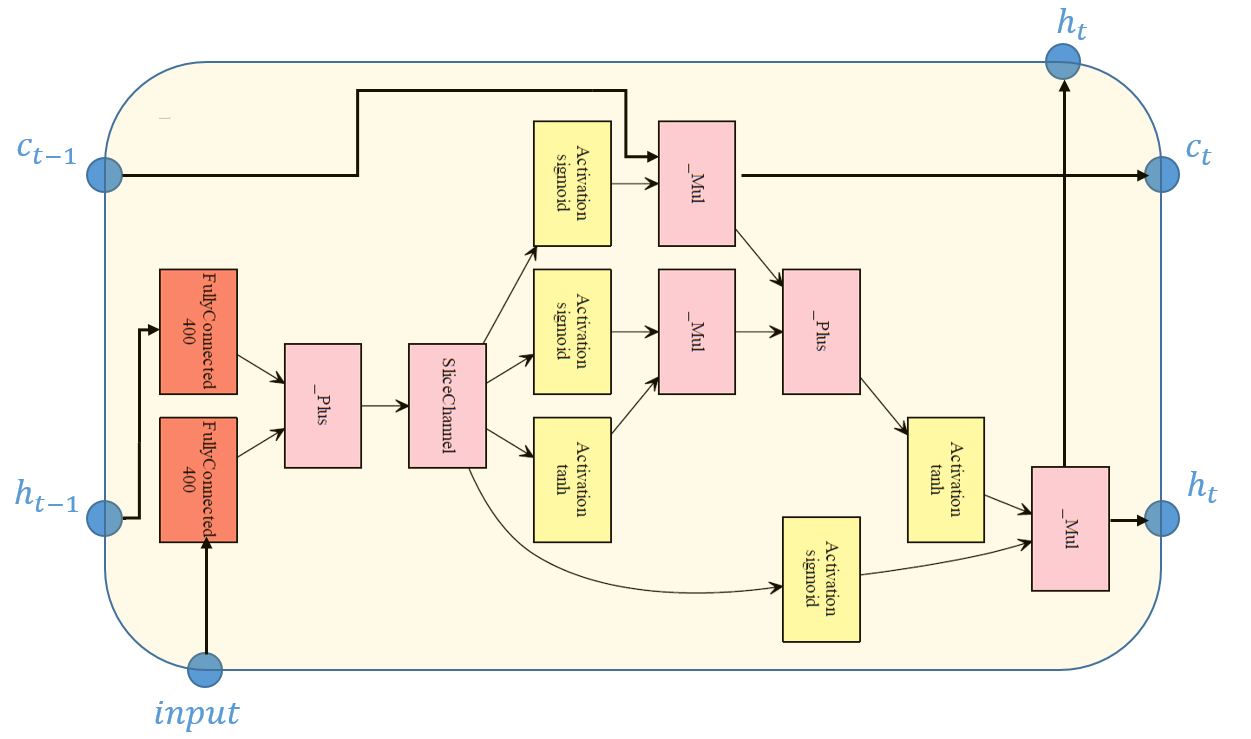}
        }
     }
     \caption{LSTM network structure (zoom to view details).}
\label{fig:lstm}
\end{figure}

\begin{figure}[h]
    \centering
        \includegraphics[width=0.95\columnwidth]{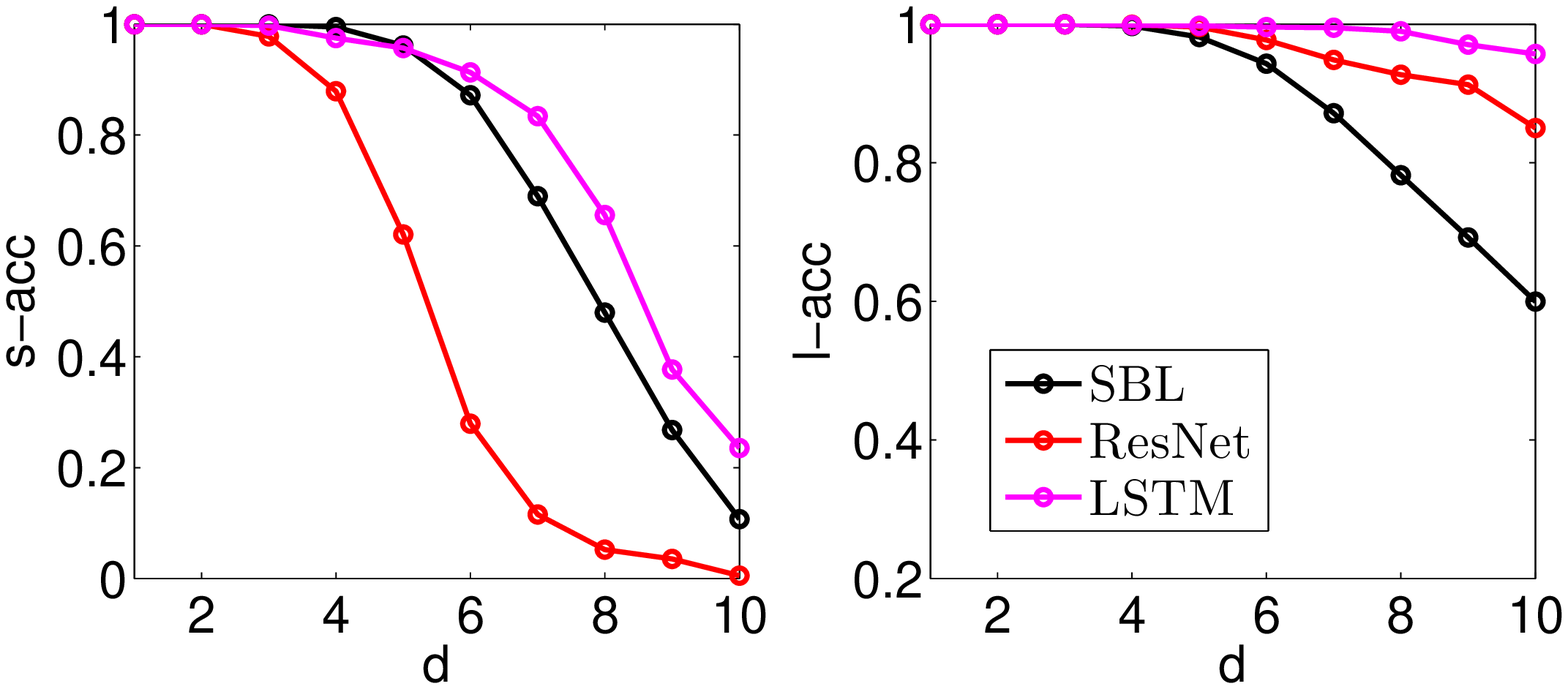}
     \caption{Sparse recovery accuracy of LSTM-Net compared with a residual network and the SBL algorithm. \emph{Left}: strict accuracy, \emph{Right}: loose accuracy.}
\label{fig:testlstm}
\end{figure}

Using a cell design from \citep{hochreiter1997long}, we adopt a two-layer LSTM network with a fixed size of 11 steps.  Figure \ref{fig:lstm} presents the specific structure.  We compare this network against both our original residual net implementation and SBL.  The later was chosen because it represents an algorithm explicitly designed to handle dictionary correlations \citep{Wipf12}. For training and testing we use the protocol from Section \ref{ssec:basic}, but with nonzero elements of $\bx^*$ having unit magnitudes.  This modification was introduced because it represents a challenging scenario for many traditional sparse optimization algorithms for technical reasons related to local minima detailed in \citep{Wipf11}.  Results are shown in Figure \ref{fig:testlstm}, where we observe that the LSTM net is even able to outperform SBL.  Further experiments and connecting analyses will be presented in future work for space considerations.

\section{Conclusions} \label{sec:discussion}

There is a clear relationship between iterative optimization rules promoting sparsity (such as those from IHT) and the layers of deep networks.  Building from this perspective, in this paper we have shown that deep networks with hand-crafted, multi-resolution structure can provably solve certain specific classes of sparse recovery problems where existing algorithms fail.  However, much like CNN-based features can often outperform SIFT on many computer vision tasks, we argue that a discriminative approach can outperform manual structuring of layers/iterations and compensate for dictionary coherence under more general conditions.  We also believe that many of the underlying principles explored herein also transfer to other applications operating at the boundary between optimization- and learning-based systems.

There is of course one important caveat with the pursuit of maximal sparsity using a deep network.  Sparse estimation problems can be partitioned into two different categories, centered around whether or not the dictionary $\bPhi$ is reusable.  In brief, learning-based methods are only feasible when a fixed (or similar) dictionary can be repeatedly used to represent a signal class of interest.  Examples include outlier removal \citep{Candes05,Ikehata12}, compressive sensing \citep{Donoho06b}, and source localization \citep{Baillet01,Malioutov05} applications where, once learned, a deep model can produce maximally sparse representations as new input signals arrive.  In contrast, other influential domains such as subspace clustering \citep{Elhamifar13} effectively require solving sparse recovery problems with a novel dictionary at each instance such that any attempt to construct a viable training set would be infeasible.  Hence optimization-based sparse estimation nonetheless remains an important tool regardless of how effective learned models can sometimes be.

\section*{Appendix}

Here we include technical proofs of our main results.  In places we rely on three standard asymptotic notations describing the order of an arbitrary function $f(x)$:
\begin{eqnarray}
f(x)=\mathcal{O}(g(x)) & \iff & \exists c>0, |f(x)|\leq c|g(x)|, \nonumber \\
f(x)=\Omega(g(x)) & \iff & \exists c>0, |f(x)|\geq c|g(x)|, \\
f(x)=\Theta(g(x)) & \iff & \exists c_1,c_2>0, c_1|g(x)|\leq |f(x)|\leq c_2|g(x)|. \nonumber
\end{eqnarray}

\subsection{Proof of Proposition \ref{prop:constrained_IHT_layer}}

Consider some $\bx^*$ where $x_1^* = 1$ and $x_i^* = \epsilon$ for $i \in \Omega$, with $|\Omega| = k-1$.  In this restricted setting, it follows that
\begin{equation}
\Psi \bx^* = \bpsi_1 + \mathcal{O}(\epsilon)   ~~ \mbox{and} ~~ \bGamma \by = \bGamma [\bphi_1 + \mathcal{O}(\epsilon)] = \bGamma \bphi_1 + \mathcal{O}(\epsilon)
\end{equation}
and therefore
\begin{equation}
\Psi \bx^* + \bGamma \by = \bpsi_1 + \bGamma \bphi_1 + \mathcal{O}(\epsilon).
\end{equation}
To ensure that $\bx^* = H_k \left(\Psi \bx^* + \bGamma \by \right)$, the largest (in magnitude) $k$ elements of $\bz \triangleq \bpsi_1 + \bGamma \bphi_1 + \mathcal{O}(\epsilon)$ must align with  $\{1, \Omega\}$, and $z_i = x_i^*$ for all $i \in \{1,\Omega\}$.  Together these conditions are necessary to ensure that the $H_k$ operator will produce $\bx^*$; however, they also imply that $\bz = \bolde_1 + \mathcal{O}(\epsilon)$, where $\bolde_1$ is a zero vector with a `1' in the first position, since no element in the complement of $\{1,\Omega\}$ can be larger than $\mathcal{O}(\epsilon)$.  Therefore we require that
\begin{equation}
\bpsi_1 + \bGamma \bphi_1 = \bolde_1 + \mathcal{O}(\epsilon).
\end{equation}
Of course since this must hold for any $\epsilon > 0$, it must be that $\bpsi_1 + \bGamma \bphi_1 = \bolde_1$.  Repeating this procedure with $x_i^* = 1$ for all $i \in \{1,\ldots,m\}$ then leads to the requirement that $\bPsi + \bGamma \bPhi = \bI$.  \hfill $\square$

\subsection{Proof of Proposition \ref{prop:constant_layer_improvement}}

With $\bGamma = \bD \bPhi^{\top} \bW \bW^{\top}$ and the stipulated rescaled dictionary, (\ref{eq:constrained_IHT_layer}) becomes
\begin{equation} \label{eq:constrained_IHT_layer2}
\bx^{(t+1)} = H_k \left[ \left(\bI -\bD \bPhi^{\top} \bW \bW^{\top} \bPhi \bD \right) \bx^{(t)} + \bGamma \by \right].
\end{equation}
For the moment, assume that $\bW$ is invertible.  Then the iteration (\ref{eq:constrained_IHT_layer2}) is consistent with the modified objective
\begin{equation} \label{eq:basic_IHT_problem2}
\min_{\bx} \tfrac{1}{2} \| \bW \by - \bW \bPhi \bD \bx \|_2^2    ~~~ \mbox{s.t. } \| \bx \|_0 \leq k.
\end{equation}
Since $\| \bx^* \|_0 \leq k$ with $\by = \bPhi \bx^*$, it follows that $\| \widetilde{\bx}^* \|_0 \leq k$ and $\widetilde{\by} = \widetilde{\bPhi} \widetilde{\bx}^*$, where $\widetilde{\bx}^* \triangleq \bD^{-1} \bx^*$, $\widetilde{\by} \triangleq \bW \by$, and $\widetilde{\bPhi} \triangleq \bW \bPhi \bD$.  We may then apply Theorem 5 from \citep{Blumensath09} using this revised system and conclude that
\begin{equation}
\|\widetilde{\bx}^{(t)} - \widetilde{\bx}^*\|_2 \leq 2^{-t}\|\widetilde{\bx}^*\|_2,
\end{equation}
which leads to the stated result.

\subsection{Proof of Corollary \ref{cor:constant_layer_improvement}}

Consider some projection operator $\bP = \bW^{\top} \bW$ onto $\mbox{null}[\bDelta_r^{\top}]$, using $\bW$ formed with $r$ orthonormal rows, which equivalently projects onto the orthogonal complement of $\mbox{range}[\bDelta_r]$.  Therefore
\begin{equation}
\bW \bPhi \bD = \bW \left[ \epsilon \bA + \bDelta_r \right] \bN \bD = \bW  \epsilon \bA
\end{equation}
 when $\bD = (\epsilon \bN)^{-1}$.  Given that elements of $\bA$ are drawn iid from $\mathcal{N}(0,1/\sqrt{n})$ and $\bW$ has orthonormal rows, elements of $\bW \bA \in \mathbb{R}^{(n-r) \times m}$ will also have iid elements from the same distribution.  Hence the stated selections for $\bW$ and $\bD$ allow us to obtain the worst-case upper bound (in expectation) found in the corollary.

\subsection{Proof of Proposition \ref{prop:existence1}}

Let $\mathcal{I}_j$ denote the set of column indeces of $\bPhi$ associated with $\bPhi_j$ and $\bv$ the vectorized concatenation of all $\bv_j$, i.e., $\bv \triangleq [\bv_1^{\top}, \ldots, \bv_c^{\top}]^{\top}$.  We then introduce the following intermediate result.

\begin{lemma}
By construction of the clustered dictionary model
\begin{equation}
\by = \bU \bz^* + \bnu,
\end{equation}
where $\bz^* \in \mathbb{R}^{c}$ is a sparse vector such that $z_j^* = \sum_{i \in \mathcal{I}_j} v_i x_i^*$ (which implies that $\|\bz^* \|_0 \leq k_c$), and $\bnu = \mathcal{O}(\epsilon)$.
\end{lemma}
\textbf{\emph{Proof:}}  Without loss of generality, we note that assuming $i \in \mathcal{I}_j$, then $n_{ii} = v_i^{-1} + \mathcal{O}(\epsilon) $.  To see this, we note that
\begin{equation}
n_{ii}^{-1} \|\bphi_i \|_2 = \| v_i  \bu_j   + \epsilon \ba_i \|_2 \leq  v_i   + \epsilon   = v_i   + \mathcal{O}(\epsilon)
\end{equation}
via the triangle inequality and the fact that $\| \bu_j \|_2 = \| \ba_i \|_2 = 1$ by assumption. Therefore the normalization constant satisfies
\begin{equation}
n_{ii} = \frac{1}{ v_i   + \mathcal{O}(\epsilon) } = \frac{1}{ v_i   } + \frac{\mathcal{O}(\epsilon)}{v_i^2   + v_i  \mathcal{O}(\epsilon) } = \frac{1}{ v_i   } + \mathcal{O}(\epsilon).
\end{equation}
We then have
\begin{equation}
\bphi_i = n_i ( v_i  \bu_j   + \epsilon \ba_i ) = \bu_j + \mathcal{O}(\epsilon)
\end{equation}
and therefore $\bphi_i x^*_i = \bu_j  x_i^* + \mathcal{O}(\epsilon)$ which naturally leads to
\begin{equation}
\bPhi \bx^* = \sum_j \sum_{i \in \mathcal{I}_j} \bu_j  x_i^* + \mathcal{O}(\epsilon)  = \sum_j \bu_j  \sum_{i \in \mathcal{I}_j} x_i^* + \mathcal{O}(\epsilon),
\end{equation}
producing the stated result.
\hfill $\square$
\vspace*{0.5cm}

This design motivates developing initial iterations based solely around detecting the correct cluster support.  To accomplish this we use $\bPsi^{(t)} = \bI - \bU^{\top} \bU$, $\bGamma^{(t)} = \bU^{\top}$, $k^{(t)} = k_c$, and $\Omega_{\tiny \mbox{on}}^{(t)} = \Omega_{\tiny \mbox{off}}^{(t)}=  \emptyset$ for the initial A-IHT iterations, which is equivalent to the standard IHT updates applied using a sparsity level of $k_c$ and a dictionary formed only with $\bU$.

Here we are effectively not requiring that $\bx^{(t)} = \bz^{(t)}$ be the same dimension at each step, or equivalently, we do not require that each $\bPsi^{(t)}$ be a square matrix (this simplifies the exposition although we could nonetheless provide an alternative argument with a fixed dimension).  Note that since $k_x \geq k_c$, then
\begin{equation}
\delta_{3k_c}\left( \bU \right) \leq \delta_{(3k_x + k_c)}\left( \bU \right) \leq \delta_{(3k_x + k_c)}\left([\bU, {\bA}(\mathcal{J}) ]\right) < 1/\sqrt{32}
\end{equation}
for any $\mathcal{J} \subset \{1,\ldots,c\}$.  This allows us to apply Theorem 5 from \citep{Blumensath09}, from which we can infer that after at most
\begin{equation} \label{eq:needed_iterations}
\tau = \log_2 \left(\frac{\| \bz^* \|_2}{\| \bnu \|_2} \right)
\end{equation}
iterations, $\bz^*$ will be estimated with accuracy
\begin{equation} \label{eq:inequality1}
\| \bz^{(\tau)} - \bz^* \|_2 \leq 6 \| \bnu \|_2 = \mathcal{O}(\epsilon).
\end{equation}
Define $z^*_{\tiny \mbox{min}}$ as the nonzero element of $\bz^*$ with smallest magnitude and suppose that $\mbox{supp}\left[ \bz^{(\tau)} \right] \neq \mbox{supp}\left[ \bz^* \right]$.   Then in order to satisfy (\ref{eq:inequality1}), it must be that $z^*_{\tiny \mbox{min}} = \mathcal{O}(\epsilon)$.  However, $z^*_{\tiny \mbox{min}}$ is independent of $\epsilon$,  the latter of which can be made arbitrarily small when the upper bound $\epsilon'$ is small leading to a contradiction.  Consequently, it must be that $\mbox{supp}\left[ \bz^{(t)} \right] = \mbox{supp}\left[ \bz^* \right]$.  Therefore we may conclude that after a finite number of iterations, we have extracted the correct cluster support, or the correct low-resolution approximation.  Of course the correct support will likely be converged to long before we reach the iteration number from (\ref{eq:needed_iterations}), but this is a worst case bound.

Of course from a practical standpoint we will not have access to $\bz^*$ such that $\| \bz^{(\tau)} - \bz^* \|_2$ is computable.  However, it can be shown that $\|\by - \bU \bz \|_2 = \Omega(z^*_{\tiny \mbox{min}})$ for any feasible $\bz$ with $\| \bz \|_0 \leq k_c$ and $\mbox{supp}[\bz] \neq \mbox{supp}[\bz^*]$.  Therefore we can monitor this observable error metric as a proxy, and when it reaches some $\mathcal{O}(\epsilon) < \Theta(z^*_{\tiny \mbox{min}}) \leq \Omega(z^*_{\tiny \mbox{min}})$ for $\epsilon$ sufficiently small, we will be guaranteed that the correct support has been found.

We now turn our attention to extracting a final estimate of $\mbox{supp}[\bx^*]$.  First we pad $\bz^{(\tau)}$ with zeros to form the vector $\bz^{(\tau+1)} = [\bz^{(\tau)} ; {\bf 0} ] \in \mathbb{R}^{m+c}$.  This is trivially accomplished using a single layer with $\bGamma^{(\tau)} = {\bf O}$, $\bPsi^{(\tau)} = [\bI ; {\bf 0}]$, and $\Omega = \{1,\ldots,c\}$.  We next adopt the filters
\begin{eqnarray}
\bPsi^{(\tau + t)} & = & \bI - [\bU, \bA ]^{\top} [\bU, \bA ] \nonumber \\
\bGamma^{(\tau + t)} & = & [\bU, \bA ]^{\top},
\end{eqnarray}
which conform with IHT updates using the dictionary $[\bU, \bA ]$ and the revised observation model $\by = [\bU, \bA ] \bz$, where we know by assumption there exists an exact solution $\bz'$ such that $\| \bz' \|_0 \leq k_x + k_c$, with $k_c$ nonzero coefficients corresponding with columns of $\bU$, and $k_x$ nonzero coefficients corresponding with columns of $\bA$.

The support templates $\Omega_{\tiny \mbox{on}}^{(t)}$ and $\Omega_{\tiny \mbox{off}}^{(t)}$ are constructed as follows.  Define $\mathcal{J}_c \triangleq \mbox{supp}\left[ \bz^{(\tau)} \right]$.  We then use
\begin{eqnarray} \label{eq:support_patterns}
\Omega_{\tiny \mbox{on}}^{(\tau + t)} & = & \mathcal{J}_c \\
\Omega_{\tiny \mbox{off}}^{(t)} & = & c + \mbox{supp}\left[ {\bA}(\{1,\ldots,c\} \backslash \mathcal{J}_c )\right]. \nonumber
\end{eqnarray}
Conceptually these selections are quite straightforward, any notational obfuscation notwithstanding.   These support patterns should be viewed in light of the new, implicit dictionary $[\bU, \bA ]$.  With $\Omega_{\tiny \mbox{on}}^{(\tau + t)}$, we are simply selecting the basis vectors associated with $\bU$ that conform with true cluster centers.  Likewise for $\Omega_{\tiny \mbox{off}}^{(t)}$ we are effectively pruning all ${\bA}(\{1,\ldots,c\} \backslash \mathcal{J}_c )$, meaning cluster details associated with clusters that have already been pruned (and the additive factor of $c$ in (\ref{eq:support_patterns}) is merely added to account for the pre-padding with $\bU$ in $[\bU, \bA ]$).

Given these selections, we are effectively running IHT with the collapsed dictionary $[\bU, {\bA}(\mathcal{J}_c ) ]$, where we already know the true sparsity profile associated with columns of $\bU$.  This corresponds with the problem of partial support recovery for nonzeros outside of $\Omega_{\tiny \mbox{on}}^{(\tau + t)}$.  Additionally, with respect to $[\bU, {\bA}(\mathcal{J}_c ) ]$ (and all zeros elsewhere), $\bz'$ represents the maximally sparse feasible solution.   This is because
\begin{equation}
\delta_{(2[k_x+k_c])}\left([\bU, {\bA}(\mathcal{J}_c ) ] \right) \leq \delta_{(3k_x + k_c)}\left([\bU, {\bA}(\mathcal{J}_c ) ] \right) <  1/\sqrt{32},
\end{equation}
and all that is required for a unique, maximally sparsity with  $k_x+k_c$ nonzeros is the much weaker inequality $\delta_{(2[k_x+k_c])}\left([\bU, {\bA}(\mathcal{J}_c ) ] \right) < 1$ \citep{Candes06}.

Known partial support allows us to loosen the requirement for guaranteed support recovery.  In particular, using modifications of Theorem 1 from \citep{Carrillo11} and the fact that $\delta_{(3k_x + k_c)}\left([\bU, {\bA}(\mathcal{J}_c ) ] \right) <   1/\sqrt{32}$, it follows that after $\tau + t$ iterations, the A-IHT reconstruction error is bounded by
\begin{equation}
\| \bz' - \bz^{(\tau + t)} \|2 \leq 2^{-t} \|\bz' - \bz^{(\tau + 1)}\|_2.
\end{equation}
Consequently, after a finite number of iterations $\bz^{(\tau + t)}$ must have matching support as $\bz'$, and moreover, as $t \rightarrow \infty$, $\bz^{(\tau + t)} \rightarrow \bz'$.  These can be mapped directly to the optimal support of $\bx^*$, from which $\bx^*$ itself is also trivially obtained.

\vskip 0.2in
\bibliography{wipf_refs}

\end{document}